\pdfoutput=1

\documentclass[11pt]{article}

\usepackage{acl}

\usepackage{times}
\usepackage{latexsym}

\usepackage[T1]{fontenc}

\usepackage[utf8]{inputenc}

\usepackage{microtype}

\usepackage{amsmath}
\usepackage{multirow}
\usepackage{listings}
\usepackage{color}
\usepackage{graphicx}
\usepackage{subcaption}
\usepackage{tcolorbox}
\usepackage{makecell}
\usepackage{booktabs}
\usepackage{cleveref}
\usepackage{colortbl}
\usepackage{tablefootnote}

\definecolor{dkgreen}{rgb}{0,0.6,0}
\definecolor{gray}{rgb}{0.5,0.5,0.5}
\definecolor{mauve}{rgb}{0.58,0,0.82}

\definecolor{dgreen}{rgb}{0.412,0.741,0.271}
\definecolor{dblue}{rgb}{0.220,0.325,0.639}
\definecolor{dred}{rgb}{0.933,0.122,0.137}

\definecolor{g1}{HTML}{b3e2cd}
\definecolor{r1}{HTML}{fdcdac}
\definecolor{w1}{HTML}{cbd5e8}
\definecolor{b1}{HTML}{fff7bc}

\definecolor{lr}{HTML}{bebada}
\definecolor{fr}{HTML}{fccde5}

\newcommand{\ie}{\textit{i}.\textit{e}.\,}
\newcommand{\eg}{\textit{e}.\textit{g}.\,}

\newcommand{\lone}[2]{\multicolumn{1}{>{\columncolor{#1}}l}{#2}}
\newcommand{\loneg}[1]{\lone{g1}{#1}}
\newcommand{\loner}[1]{\lone{r1}{#1}}

\newcommand{\loneb}[1]{\lone{b1}{#1}}
\newcommand{\lonelr}[1]{\lone{lr}{#1}}
\newcommand{\lonefr}[1]{\lone{fr}{#1}}

\newcommand{\ltwo}[2]{\multicolumn{2}{>{\columncolor{#1}}l}{#2}}
\newcommand{\ltwog}[1]{\ltwo{g1}{#1}}
\newcommand{\ltwor}[1]{\ltwo{r1}{#1}}

\newcommand{\ltwob}[1]{\ltwo{b1}{#1}}

\newcommand\blfootnote[1]{%
  \begingroup
  \renewcommand\thefootnote{}\footnote{#1}%
  \addtocounter{footnote}{-1}%
  \endgroup
}

\title{Feeding What You Need by Understanding What You Learned}

\author{
    Xiaoqiang Wang\textsuperscript{\rm 1$*$},
    Bang Liu\textsuperscript{\rm 2$*$$\dagger$},
    Fangli Xu\textsuperscript{\rm 3},
    Bo Long\textsuperscript{\rm 4},
    Siliang Tang\textsuperscript{\rm 1$\ddagger$} \and
    Lingfei Wu\textsuperscript{\rm 5$\ddagger$} \\
    \textsuperscript{\rm 1}Zhejiang University, \textsuperscript{\rm 2}Universit{\'e} de Montr{\'e}al \& Mila, \textsuperscript{\rm 3}Squirrel AI Learning \\
    \textsuperscript{\rm 4}JD.COM, \textsuperscript{\rm 5}JD.COM Silicon Valley Research Center \\
    \{\texttt{xq.wang, siliang}\}\texttt{@zju.edu.cn} \\
    \texttt{bang.liu@umontreal.ca, lili@yixue.us} \\
    \{\texttt{bo.long, lingfei.wu}\}\texttt{@jd.com} \\
}

\begin{document}
\maketitle
\begin{abstract}
    Machine Reading Comprehension (MRC) reveals the ability to understand a given text passage and answer questions based on it.
    Existing research works in MRC rely heavily on large-size models and corpus to improve the performance evaluated by metrics such as Exact Match ($EM$) and $F_1$.
    However, such a paradigm lacks sufficient interpretation to model capability and can not efficiently train a model with a large corpus.
    In this paper, we argue that a deep understanding of model capabilities and data properties can help us feed a model with appropriate training data based on its learning status.
    Specifically, we design an MRC capability assessment framework that assesses model capabilities in an explainable and multi-dimensional manner.
    Based on it, we further uncover and disentangle the connections between various data properties and model performance.
    Finally, to verify the effectiveness of the proposed MRC capability assessment framework, we incorporate it into a curriculum learning pipeline and devise a Capability Boundary Breakthrough Curriculum (CBBC) strategy, which performs a model capability-based training to maximize the data value and improve training efficiency.
    Extensive experiments demonstrate that our approach significantly improves performance, achieving up to an 11.22\% / 8.71\% improvement of $EM$ / $F_1$ on MRC tasks.
\end{abstract}

\blfootnote{$^*$Equal contribution.}
\blfootnote{$^\dagger$Canada CIFAR AI Chair.}
\blfootnote{$^\ddagger$Corresponding authors.}

\begin{table*}[h!]
    \resizebox{1.0\textwidth}{!}{
        \begin{tabular}{llll}
            \toprule
            \multicolumn{1}{c|}{\textbf{Capability $c_i$}} & \multicolumn{2}{c|}{\textbf{Subclasses}}& \multicolumn{1}{c}{\textbf{Metrics $m_i^j$}} \\
            \midrule
            \multirow{2}{*}{\bfseries Reading words}& \ltwog{Recognize vocabulary}& \loneg{Intra-n~\cite{gu2018dialogwae} and Ent-n~\cite{serban2017hierarchical}.} \\
            & \ltwog{Recognize function words}& \loneg{Frequency of function words.} \\
            \hline
            \multirow{2}{*}{\bfseries Reading sentences}& \ltwor{Recognize grammaticality}& \loner{Height and width of a question's constituency parsing tree.} \\
            & \ltwor{Readability}& \loner{Readability metrics.} \\
            \hline
            \multirow{2}{*}{\makecell{\bfseries Understanding \\ \bfseries words}}& \ltwob{Arithmetic operation}& \loneb{Frequency of numerical expressions (CD tag).} \\
            & \ltwob{Logical operation}& \loneb{Frequency of logically qualified words such as \emph{any}, \emph{all} and \emph{every}.} \\
            \hline
            \multirow{9}{*}{\makecell{\bfseries Understanding \\ \bfseries sentences}}& \cellcolor{lr}& \lonelr{Syntactic and semantic overlap} & \lonelr{BLEU-n~\cite{papineni2002bleu}, BERTScore~\cite{zhang2020bertscore}}  \\
            & \cellcolor{lr}& \lonelr{}& \lonelr{and MoverScore~\cite{zhao2019moverscore} between the context and question.} \\
            & \cellcolor{lr}& \lonelr{Coreference resolution}& \lonelr{Frequency of personal and possessive pronouns, such as PRP and} \\
            & \cellcolor{lr}& \lonelr{}& \lonelr{PRP\$ tags.} \\
            & \multirow{-5}{*}{\cellcolor{lr}\makecell{Linguistic \\ reasoning}}& \lonelr{Con/Dis-junction, negation}& \lonelr{Frequency of coordinating junctions, such as \emph{and} and \emph{or}.} \\ \cline{2-4}
            & \cellcolor{fr}& \lonefr{Causality}& \lonefr{Frequency of causal clauses, such as \emph{because}  and  \emph{the reason for}.} \\
            & \cellcolor{fr}& \lonefr{Spatial/Temporal}& \lonefr{Frequency of spatial/temporal expressions, such as \emph{before}, \emph{after}.} \\
            & \cellcolor{fr}& \lonefr{}& \lonefr{and \emph{in front of}.} \\
            & \multirow{-4}{*}{\cellcolor{fr}\makecell{Factual \\ reasoning}}& \lonefr{Multi-hop reasoning}& \lonefr{Number of supporting evidences.}\\

            \bottomrule
        \end{tabular}
    }
    \caption{Example set of MRC model capabilities $\{ c_i \}$ and corresponding metrics $\{ m_i^j \}$. See Section~\ref{subsec:assessframework} for details.}
    \vspace{-5mm}
    \label{tab:capabilities}
\end{table*}

\section{Introduction}
\label{sec:introduction}
A competency assessment is used to measure someone's capabilities against the requirements of their job \cite{web:competency}. In other words, it measures how (behaviors) someone does the what (task or skill).
By showing what it looks like to be good in a job, a competency assessment can effectively empower and engage people who want to understand and improve their unique skill profile and tell them what action to take to close any gaps so they can own their development.
A natural question that arises here is: can we develop competency assessments for machine learning models to help better understand their capabilities and improve their performance on a given task?

In this paper, we focus on competency assessments for machine reading comprehension (MRC). MRC is a core task in natural language processing (NLP) that aims to teach machines to understand human languages and answer questions~\cite{zeng2020survey,chen2019graphflow}.
Recently, pre-trained language models (LMs)~\cite{mikolov2013efficient,peters2018deep,pennington2014glove,devlin_bert_2018} have demonstrated superior performance on MRC tasks by pre-training on large amounts of unlabeled corpus and fine-tuning on MRC datasets.
The performance is usually evaluated by metrics such as Exact Match ($EM$) and $F_1$ score, lacking interpretability to the capabilities of a model. That is to say, such metrics only tell how good a model performs overall on a specific dataset, but uncovers little about what specific skills a model has gained and the level of each skill.

We argue that the value of each data sample varies during the training process of a model, depending on the model's current capabilities. A deep understanding of the model's intrinsic capabilities can help us estimate each data sample's learning value and better manage the training process to improve the training efficiency.
Take student learning as an example.
There is no doubt that a college student can do well in solving primary school level exercises, but such exercises do not help improve his/her ability.
On the contrary, a primary school student can not acquire knowledge efficiently from college-level exercises due to the big gap between his/her current knowledge or skills and the requirement to solve the exercises.
We need to measure the ability of a student and then choose the appropriate exercises accordingly.

Existing works on interpreting MRC model capabilities concentrate on analyzing a model's behavior with adversarial data~\cite{jia2017adversarial}, or defining the prerequisite skills to solve a specific dataset~\cite{sugawara2017prerequisite}.
However, these works require costly human annotation efforts or ignore the fact that model capabilities change during the training progresses.


In this paper, we design a competency assessment framework for MRC model capabilities. Specifically, we define four major capability dimensions for understanding text and solving MRC tasks: \emph{reading words}, \emph{reading sentences}, \emph{understanding words} and \emph{understanding sentences}, which are inspired by the computational models of human text comprehension in psychology~\cite{kintsch1988role}.
Based on the proposed framework, we can obtain a more appropriate assessment of model capabilities than the regular $EM$ or $F_1$ metrics. 

Furthermore, we analyze a variety of data properties to estimate how good a model has to be to solve a specific data sample and identify the relationships between data properties and model performance. This greatly helps us estimate the learning value of each training sample.
Based on this analysis, we discover a very common situation: if a sample is scored as a high value in one capability dimension, the other dimensions have the same tendency as well, and vice verse.
To alleviate these inevitable correlations, we utilize data whitening to quantify each sample as four capability-specific scores in a decorrelated fashion.


Finally, to reveal the potential usefulness of our proposed competency assessment framework and evaluate its efficiency, we employ it in a curriculum learning pipeline and design a Capability Boundary Breakthrough Curriculum (CBBC) strategy.
This strategy gradually enlarges the model capability boundary by picking samples around the boundary and breaking through it.
Based on the analysis of model capabilities and data properties, we feed the model with training samples that are neither too simple nor too hard for it to solve.
Extensive experiments on four benchmark datasets demonstrate that our approach significantly improves the performance of existing MRC models, achieving up to an 11.22\% / 8.71\% improvement of $EM$ / $F_1$ on MRC tasks.
These results show the reasonability and effectiveness of our proposed assessment framework and provide a widely applicable measurement for dealing with the relationship between the model capability and data quality.

\section{Competency Assessment of MRC Capabilities}
\label{sec:capability-evaluation}
In this section, we first formulate our competency assessment framework of 4-dimensional MRC capabilities.
Based on this framework, the data properties related to each capability dimension are described as corresponding heuristic metrics.
We then uncover the relationship between various data properties and model performance in a decorrelated manner, quantifying each sample as 4-dimensional capability-specific scores with little correlation. 

\subsection{Assessment Framework Formulation}
\label{subsec:assessframework}
Human text comprehension has been studied in psychology for a long time.
Constructionist, landscape model and computational architectures have been proposed for such comprehension~\cite{mcnamara2009toward}.
Among them, the construction-integration (CI) model~\cite{kintsch1988role} is one of
the most basic and influential theories.
The CI model assumes three different representation levels (surface structure, textbase and situation model) and a two-step process (construction and integration) to understand text comprehensively.
It first constructs the propositions (\ie textbase) from the raw textual input (\ie surface structure), then integrates the local connections into a globally coherent representation (\ie situation model).
Based on this situation model, a given text is understood comprehensively and can even be grounded to other modalities.
Inspired by the two-step process of the CI model, we formulate our assessment framework by 4-dimensional capabilities as summarized in Table~\ref{tab:capabilities}.
We sketch out the meaning of each MRC capability $\{ c_i \}_{i=1}^{4}$ and highlight some heuristic metrics $\{ m_i^j \}_{j=1}^{n(i)}$ (where $n(i)$ means the number of metrics to measure a sample's learning value to capability $c_i$) as follows.

\begin{figure}[!t]
\begin{tcolorbox}
    \small
    \textbf{Context:} \textcolor{dgreen}{James is a trouble making Turtle }. One day, James went to the grocery store and pulled all the pudding off the shelves and ate two jars. \textcolor{dblue}{Then} he \textcolor{dred}{walked to} the fast food restaurant and ordedred 15 bags of fries.
    
    \textbf{Q1:} Who \textcolor{dgreen}{is the trouble making turtle}?

    \textbf{A1:} James

    \textbf{Q2:} Where did James \textcolor{dred}{go} \textcolor{dblue}{after} he went to the grocery store?

    \textbf{A2:} A fast food restaurant

    \textbf{Requidred capabilities:} \textcolor{dgreen}{syntactic matching}, \textcolor{dblue}{temporal relation}, \textcolor{dred}{semantic overlap}
\end{tcolorbox}
\caption{Two example questions $Q1$ and $Q2$ with different difficulties require different capabilities. }
\vspace{-5mm}
\label{fig:intro-examples}
\end{figure}

\noindent
\textbf{Reading words.} \
To formulate the surface structure of the CI model in our framework, we first highlight the text representation at the verbal or linguistic level.
Theoretically, the units at the linguistic level are the words that make up the text and the hierarchical sentence constituents to which these words belong.
Empirically, \citet{sugawara2018makes} has shown that some questions are answered correctly by just reading the first $k$ tokens.
Similarly, the perturbation-based experiments of \citet{nema-khapra-2018-towards} have demonstrated the significant influence of four types of words (\ie content words, named entities, question types, and function words) on an MRC question.
Therefore, the dimension of \emph{reading words} is defined as recognizing the observed vocabulary and the special words' appearance (\ie function words). 
In this study, The former is implemented as \emph{Intra-n}~\cite{gu2018dialogwae} and \emph{Ent-n}~\cite{serban2017hierarchical} to measure vocabulary distribution, while the latter is computed as the frequency of corresponding words.

\noindent
\textbf{Reading sentences.} \
The rules that are used to form a sentence using the aforementioned linguistic units are conventional phrase-structure grammars.
Consequently, before understanding the information contained in a text, an MRC system inevitably requires capturing the sentence structure and handling the possible obscure words. 
We define the dimension of \emph{reading sentences} as recognizing grammaticality and readability, and they are implemented by constituency parsing tree statistics and readability metrics\footnote{https://py-readability-metrics.readthedocs.io/}, respectively.

\noindent
\textbf{Understanding words.} \
The semantic representation of text is usually established by local and global links according to the linguistic units at word-level and sentence-level, respectively.
To reflect the local semantic structure, we design the dimension of \emph{understanding words} to assess how well an MRC model understands the relationships between words.
In this work, we exemplify two relations (\ie the arithmetic operations and logical items) that usually have salient patterns in the text.
The former directly focuses on statistical and operational reasoning from the text, while the latter deals with the reasoning of predicate logic, \eg conditionals and qualifiers.
Inspired by the human annotation process~\cite{boratko2018systematic, schlegel2020framework}, where the annotators are asked to label as many reasoning skills as possible by paying more attention to corresponding indicative words, the sub-capabilities of this dimension is quantified as the frequency of those words.

\noindent
\textbf{Understanding sentences.} \
Integrating the local structures into a global representation requires not only the text itself but also specific knowledge.
To simplify the forms of knowledge, we divide the dimension of \emph{understanding sentences} into two subclasses, linguistic and factual reasoning.
They respectively mean understanding the relationship between sentences based on linguistics and the events (\ie five dimensions including time, space, causation, intentionality, and objects).
Among metrics of this dimension, \emph{BERTScore}~\cite{zhang2020bertscore}, \emph{MoverScore}~\cite{zhao2019moverscore} and \emph{LS\_score}~\cite{wu2020unsupervised} are used to measure semantic overlap between the context and question and multi-hop reasoning is an extra particular subclass on the HotpotQA~\cite{yang2018hotpotqa,cheng2021guiding} dataset.
For the other sub-capabilities of this dimension, we consider lessons of the ablation operations performed by \citet{sugawara2020assessing} to observe the performance change of the MRC model and quantify them using the corresponding indicative structures.

Consider the two examples questions shown in Figure~\ref{fig:intro-examples}. To solve $Q1$, an MRC system just needs to match the words between the question and context. However, $Q2$ requires understanding temporal relations among the events (went to the grocery store $\rightarrow$ walked to the fast-food restaurant) and the verb semantics (walk to means go to). Therefore, $Q2$ is more challenging to the MRC system than $Q1$.
\emph{Please refer to Appendix~\ref{sec:examples-metrics} for more detailed examples and descriptions of our employed metrics.}

\begin{table}[!t]
    \resizebox{1.0\columnwidth}{!}{
        \begin{tabular}{c|c|cc|cc|cc|cc}
            \toprule
            \large
            & \multirow{2}{*}{\textbf{Value}}& \multicolumn{2}{c|}{\textbf{SQuADv1}}& \multicolumn{2}{c|}{\textbf{SQuADv2}}& \multicolumn{2}{c|}{\textbf{HotpotQA}}& \multicolumn{2}{c}{\textbf{RACE}} \\
            & & $r$& $p$& $r$& $p$& $r$& $p$& $r$& $p$ \\
            \midrule
            \midrule
            
            \multirow{12}{*}{\rotatebox[origin=c]{90}{\textbf{$F_1$}}}& \multirow{2}{*}{$v_1$}& -0.131& 0.000& -0.135& 0.000& -0.146& 0.007& -0.129& 0.010 \\
            & &-0.120& 0.018& -0.124& 0.026& -0.135& 0.009& -0.118& 0.017 \\
            \cline{2-10}

            & \multirow{2}{*}{$v_2$}& -0.162& 0.002& -0.152& 0.027& -0.174& 0.022& -0.173& 0.003 \\
            & &-0.154& 0.000& -0.144& 0.025& -0.166& 0.000& -0.165& 0.005 \\
            \cline{2-10}

            & \multirow{2}{*}{$v_3$}& -0.134& 0.000& -0.130& 0.000& -0.141& 0.029& -0.135& 0.011 \\
            & &-0.163& 0.016& -0.159& 0.026& -0.170& 0.006& -0.164& 0.017 \\
            \cline{2-10}

            & \multirow{6}{*}{$v_4$}& -0.166& 0.001& -0.155& 0.018& -0.182& 0.000& -0.181& 0.019 \\
            & &-0.198& 0.020& -0.187& 0.026& -0.214& 0.018& -0.213& 0.006 \\
            & &-0.208& 0.015& -0.197& 0.020& -0.224& 0.013& -0.223& 0.002 \\
            & &-0.206& 0.000& -0.195& 0.002& -0.222& 0.010& -0.221& 0.001 \\
            & &-0.168& 0.023& -0.157& 0.022& -0.184& 0.023& -0.183& 0.006 \\
            & &-0.168& 0.010& -0.157& 0.004& -0.184& 0.002& -0.183& 0.000 \\

            \hline

            \multirow{12}{*}{\rotatebox[origin=c]{90}{\textbf{$F_{logits}$}}}& \multirow{2}{*}{$v_1$}& \textbf{-0.144}& 0.012& \textbf{-0.151}& 0.000& \textbf{-0.165}& 0.007& \textbf{-0.142}& 0.010 \\
            & & \textbf{-0.133}& 0.018& \textbf{-0.140}& 0.026& \textbf{-0.154}& 0.009& \textbf{-0.131}& 0.017 \\
            \cline{2-10}

            & \multirow{2}{*}{$v_2$}& \textbf{-0.188}& 0.002& \textbf{-0.175}& 0.000& \textbf{-0.205}& 0.022& \textbf{-0.203}& 0.003 \\
            & & \textbf{-0.180}& 0.000& \textbf{-0.167}& 0.025& \textbf{-0.197}& 0.028& \textbf{-0.195}& 0.005 \\
            \cline{2-10}

            & \multirow{2}{*}{$v_3$}& \textbf{-0.163}& 0.000& \textbf{-0.157}& 0.000& \textbf{-0.172}& 0.000& \textbf{-0.163}& 0.011 \\
            & & \textbf{-0.192}& 0.016& \textbf{-0.186}& 0.026& \textbf{-0.201}& 0.006& \textbf{-0.192}& 0.017 \\
            \cline{2-10}

            & \multirow{6}{*}{$v_4$}& \textbf{-0.206}& 0.001& \textbf{-0.192}& 0.018& \textbf{-0.228}& 0.023& \textbf{-0.226}& 0.019 \\
            & & \textbf{-0.238}& 0.020& \textbf{-0.224}& 0.026& \textbf{-0.260}& 0.018& \textbf{-0.258}& 0.006 \\
            & & \textbf{-0.248}& 0.015& \textbf{-0.234}& 0.020& \textbf{-0.270}& 0.013& \textbf{-0.268}& 0.002 \\
            & & \textbf{-0.246}& 0.023& \textbf{-0.232}& 0.002& \textbf{-0.268}& 0.010& \textbf{-0.266}& 0.001 \\
            & & \textbf{-0.208}& 0.023& \textbf{-0.194}& 0.022& \textbf{-0.230}& 0.023& \textbf{-0.228}& 0.006 \\
            & & \textbf{-0.208}& 0.010& \textbf{-0.194}& 0.004& \textbf{-0.230}& 0.002& \textbf{-0.228}& 0.000 \\

            \bottomrule
        \end{tabular}
    }
    \caption{
    The Pearson’s correlation ($r$) between capability-specific value $v_i$ and model performance.
    Stronger correlations are marked in \textbf{bold}.
    All the correlations are with the p-value < 0.05.
    }
    \vspace{-5mm}
    \label{tab:analysis-correlation}
\end{table}

\vspace{-0.1mm}
\subsection{Relationship Between Data Properties and Model Performance}
\label{sec:relation-analysis}
Based on our assessment framework, the learning value of each sample is also decomposed into four dimensions, namely capability-specific values.
In this section, we first uncover the connection between the capability-specific values and model performance from four dimensions and then recalibrate the connection by removing the inter-dimension correlations.

\noindent
\textbf{Capability-specific value.} \
Given a sample $x$, we represent it by four capability-specific value (denoted as $\{ v_i(x) \}_{i=1}^{4}$) to reflect its learning value for each capability dimension.
According to our assessment framework, $v_i(x)$ can be computed by merging the corresponding metrics $\{ m_i^j(x) \}_{j=1}^{n(i)}$.
Specifically, considering the sensitivity of capability-specific value to different ranges of the metric score, we normalize each raw metric $m_i^j(x)$ from its original scale to range $[0, 1]$ by the cumulative density function (CDF) as \citet{platanios-etal-2019-competence}, which is denoted as $\widetilde{m_i^j(x)}$.
In this work, the normalization computes the cumulative density from a higher model performance to ensure that the normalized metric and model performance are negatively correlated.
The capability-specific score $v_i(x)$ is formulated as: $v_i(x) = \frac{1}{n(i)} \sum_{j=1}^{n(i)}{\widetilde{m_i^j(x)}}$.

\begin{figure}[!t]
	\centering
	
    \includegraphics[width=\columnwidth]{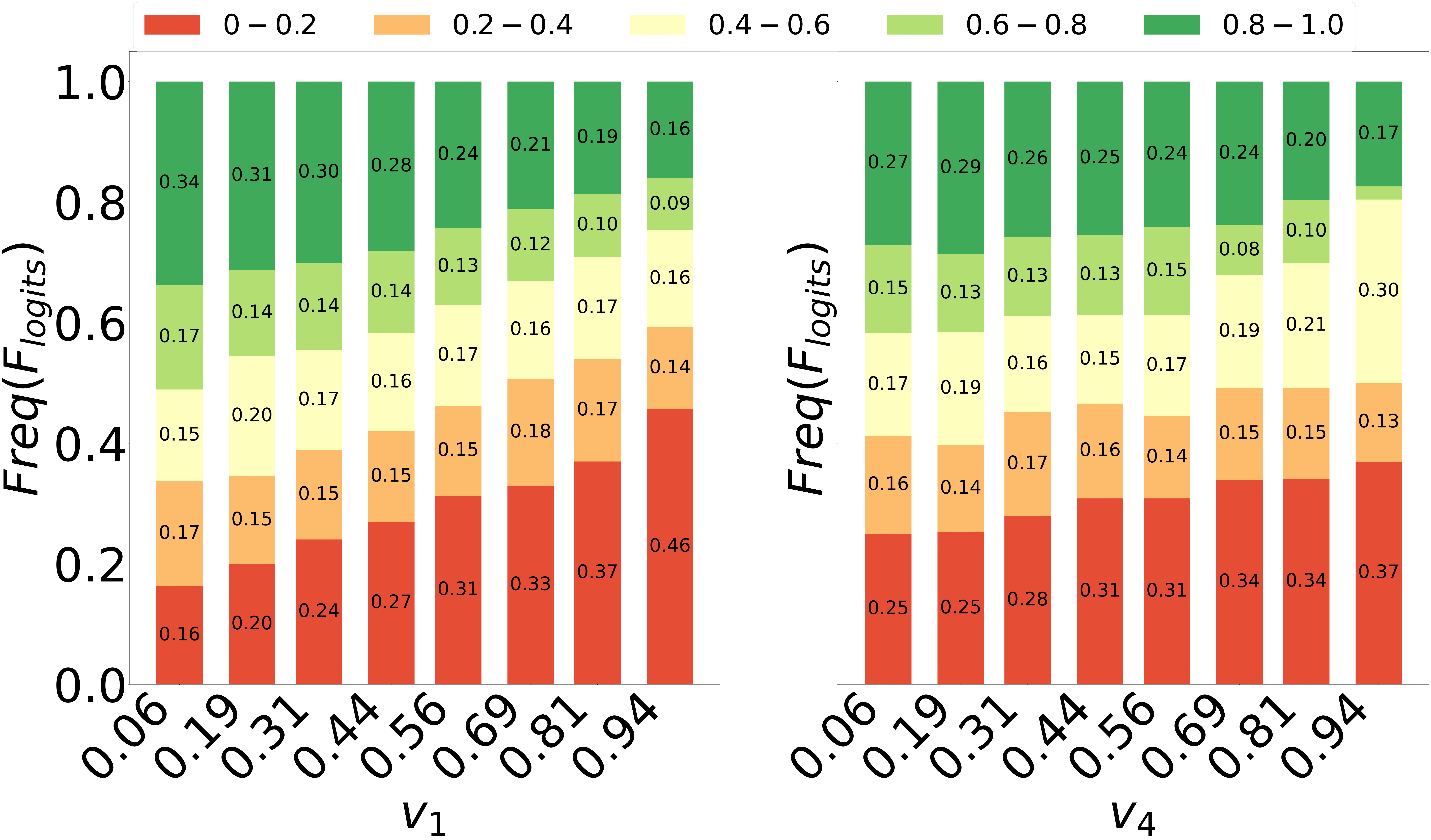}

	\caption{Bar diagram illustrating the relationship between the distribution of model performance and different ranges of $v_i$.
	Horizontal axes represent the different score ranges of $v_i$ of samples, and the vertical axis means the performance distribution by the frequency of $F_{logits}$ on five levels (plotted in five colors).
	}
	\vspace{-5mm}
	\label{fig:value-distribution}
\end{figure}

\noindent
\textbf{Analysis between capability-specific values and model performance.} \
For each sample $x$, we obtain a 4-dimensional score $\{ v_i(x) \}_{i=1}^{4}$.
It is necessary to explore the relationship between samples' $v_i(x)$ and model performance for knowing about what specific capabilities a model has gained and the level of each capability.
In this work, we employ BERT-base~\cite{devlin_bert_2018} as the MRC model and train it respectively on training split of datasets SQuADv1~\cite{rajpurkar2016squad}, SQuADv2~\cite{rajpurkar2018know}, HotpotQA~\cite{yang2018hotpotqa} and RACE~\cite{lai2017race}.
We then analyze the correlations between four capability-specific scores and the model's overall performance on the corresponding dev split.
In addition to $F_1$, we also report the results of scaled $F_1$ (denoted as $F_{logits}$) by taking the model's confidence to an answer span or candidate into account.
$F_{logits}$ is computed as:
\begin{equation}
    F_{logits} = \begin{cases}
        F_1 * ln(slog) * ln(elog) & \text{or} \\
        F_1 * ln(candlog) & 
    \end{cases}
\end{equation}
where $slog$ and $elog$ mean the model output logits for start and end token in answer extraction style questions, and $candlog$ represents the largest logits among all candidate answers.

Table~\ref{tab:analysis-correlation} quantitatively shows the Pearson’s correlation coefficients ($r$) between capability-specific values and model performance.
From the results, we have the following observations:
First, each capability-specific score has a relatively strong correlation with the model performance under a statistically significant guarantee, showing the reasonability of our capability-based assessment framework.
Second, $F_{logits}$ shows better relevancy than $F_1$, which indicates that $F_{logits}$ is a more appropriate performance measurement in our framework.

We further explore the distribution of model performance over different ranges of $v_i$.
The distribution diagrams of $v_1$ and $v_4$ are shown in Figure~\ref{fig:value-distribution}.
There are two inspiring characteristics in this diagram:
First, among all the bins of $v_i$, the frequency of prediction results within the intermediate range ($0.4\sim0.6$) are similar ($ \approx 50\%$).
Second, as the $v_i$ increases, the frequency of prediction results within a low range ($0.0\sim0.2$) also increases, while the one of a high range ($0.8\sim1.0$) decreases.
These observations reveal that the samples with high $v_i$ can be used in indicative measurements to the corresponding model capability $c_i$.
\emph{Please refer to Appendix~\ref{sec:additional-diagrams} for more diagrams illustrating this relationship.}

\begin{figure}[!t]
	\centering	
	\begin{subfigure}{0.32\columnwidth}
		\includegraphics[width=\textwidth]{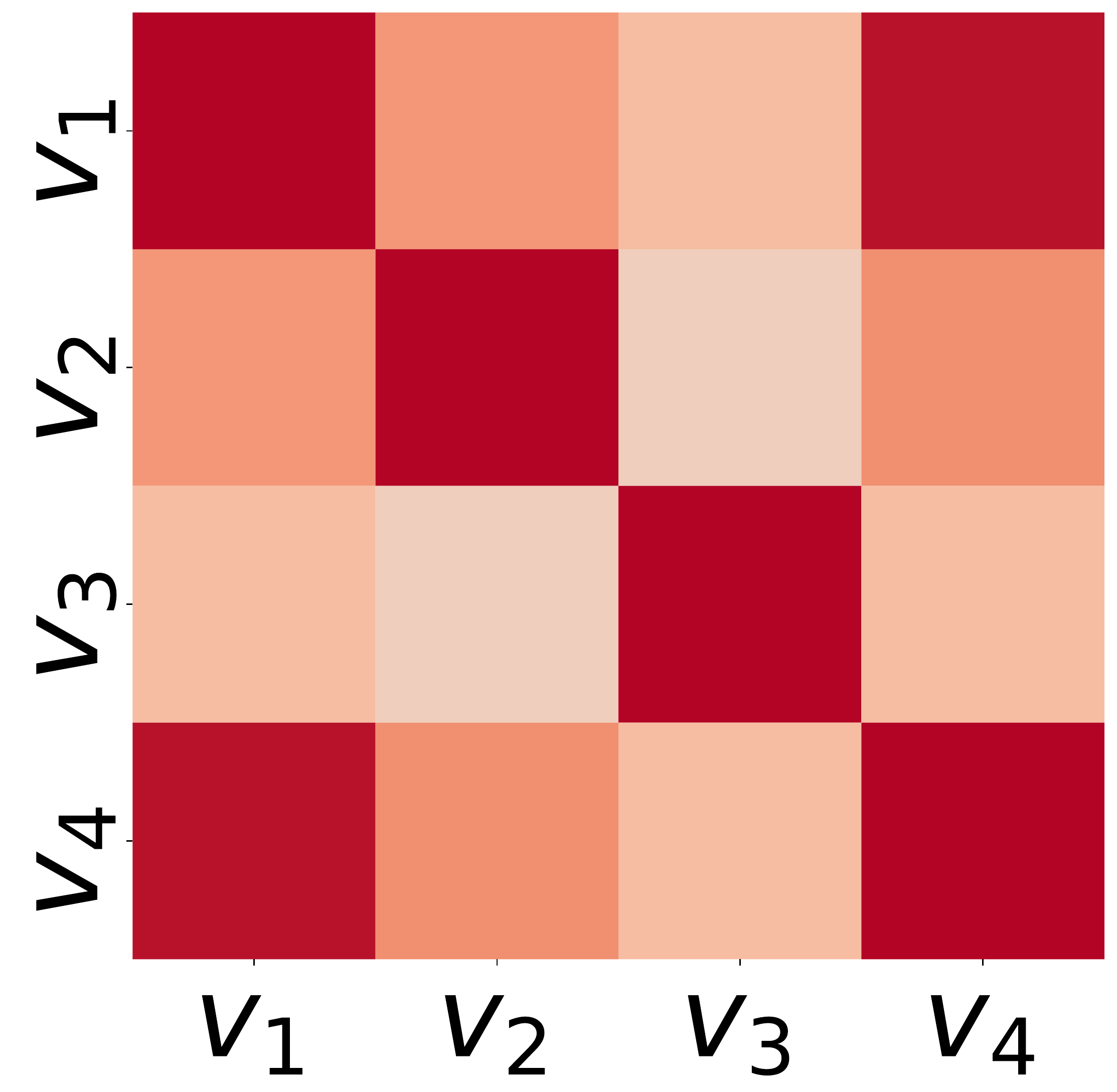}
		\subcaption{Before applying inter-dimension decorrelation.}
		\label{fig:capabilities-correlation-before}
	\end{subfigure}
	\hspace{3mm}
	\begin{subfigure}{0.425\columnwidth}
		\includegraphics[width=\textwidth]{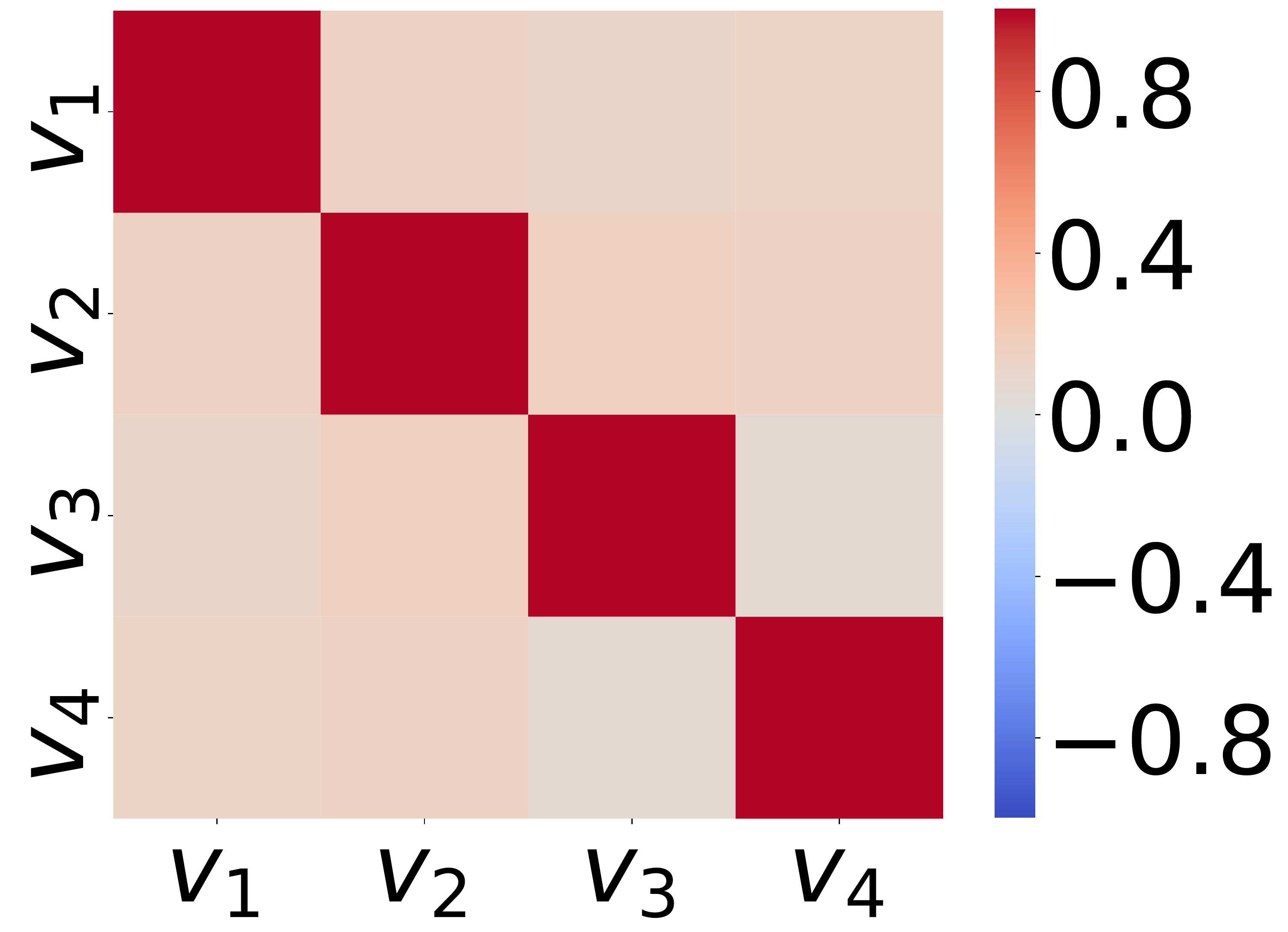}
		\subcaption{After whitening capability-specific scores using ZCA.}
		\label{fig:capabilities-correlation-after}
	\end{subfigure}

	\caption{Pairwise correlations of capability-specific scores before and after inter-dimension decorrelation.}
	\vspace{-5mm}
    \label{fig:capabilities-correlation}
\end{figure}

\noindent
\textbf{Inter-dimension decorrelation.} \
Let $\mathcal{V} = \{v_i|i = 1, \cdots, 4\}$.
Pairwise correlations of $\mathcal{V}$ are illustrated in Figure~\ref{fig:capabilities-correlation-before} in a heatmap fashion.
The results show a common situation where if a sample is difficult (scored as high capability-specific value) in a dimension, the other dimensions have the same tendency and vice versa.
To alleviate the inevitable correlations and construct a clear value representation for our following specific application scenario (\ie CBBC), we eliminate the 4-dimensional capabilities by decorrelation.
Specifically, we employ zero-phase component analysis (ZCA) whitening~\cite{bell1996edges} to diagonalize the covariance matrix while keeping the local information of the samples as much as possible.

As shown in Figure~\ref{fig:capabilities-correlation-after}, the 4-dimensional capabilities are not highly correlated after inter-dimension decorrelation, which can be in favor of constructing clear indicators for our following data sampling in CBBC.

\begin{figure}[!t]
    \includegraphics[width=\columnwidth]{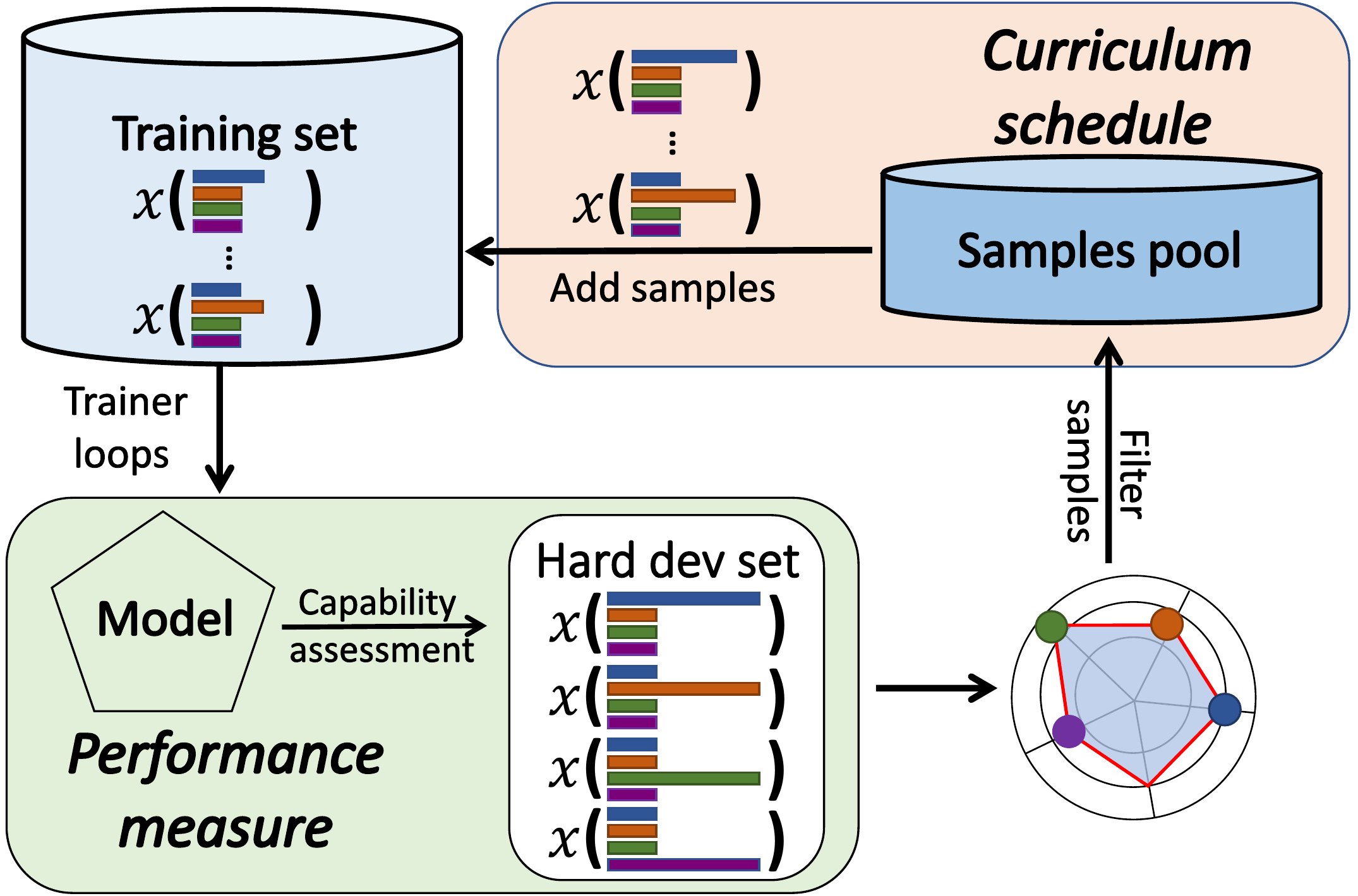}
    \caption{ Illustration of our capability boundary breakthrough curriculum learning (CBBC).
    Different capabilities and their levels are represented in different colors and the length of the color bar, respectively. A longer color bar indicates a stronger capability.
    }
    \label{fig:pipeline}
    \vspace{-5mm}
\end{figure}

\section{Improve Learning Efficiency with Competency Assessment}
\label{sec:curriculum}

In this section, our universal assessment framework of the model capability is adapted into a specific MRC training scenario to evaluate its usefulness and efficiency.
Specifically, we embed our proposed assessment framework into a curriculum learning pipeline and make a capability boundary breakthrough curriculum learning (CBBC) strategy.
Based on the assessment framework, our CBBC can guide a model to learn according to its capability boundary by understanding what the model has learned from data (\ie capability-specific value $v_i$) and choosing appropriate samples with comparable learning values from four dimensions.
It is worth noting that our competency assessment framework is also applicable to other training pipelines that balance the relationship between the model capabilities and data properties, such as active learning~\cite{settles2009active} and self-training~\cite{mihalcea2004co} (Appendix~\ref{sec:additional-pipelines}).

Figure~\ref{fig:pipeline} shows an illustration of the pipeline of our CBBC.
Following the original formulation of curriculum learning~\cite{bengio2009curriculum},
our CBBC organizes all samples by a sequence of ordered training stages $\{ s \}_{s=1}^{S}$ and corresponding training sets $\{ D^s \}_{s=1}^{S}$ with an easy-to-difficult fashion.
The classic curriculum learning works~\cite{soviany2021curriculum} usually consist of two essential components: the performance measurer and the curriculum scheduler.
In general, the measurer is used to determine the learning status of a model by evaluating performance, while the scheduler is responsible for deciding when and how to update the curriculum by selecting the input samples.

In our work, the measurer and scheduler are implemented by analyzing the multi-dimensional capability levels of the model interpretably and measuring the capability-specific values of the data in a decorrelated way, respectively.
That is to say, the only difference between our CBBC and the original curriculum learning design is incorporating MRC capability assessment into the curriculum learning.
Without significantly increasing the complexity of the pipeline, our proposed assessment framework can generally empower the MRC training pipeline in a plug-and-play manner.

\noindent
\textbf{Performance measurer.}
Recall what we have discussed in Section~\ref{sec:relation-analysis} that the samples with high $v_i$ can be used in indicative measurements to the corresponding model capability $c_i$.
In this work, we use samples scored in the top-$k$ of each capability-specific value to assess the corresponding model capability.
More precisely, we first evaluate the model on the dev set and obtain an average $F_{logits}$ for each capability on the corresponding top-$k$ subset.
Then partial correlation~\cite{baba2004partial} (denoted as $\rho_i$) between dimension $v_i$ and $F_{logits}$ is computed to mask the contributions of the other dimensions $\mathcal{V}\backslash\{v_i\}$.
After that, each model capability on stage $s$ is quantified as: $c_i^s = \frac{\rho_i}{\sum_{j=1}^{4}{\rho_j}} F_{logits}$.
Empirically, we set $k$ in top-$k$ as $32$.

\noindent
\textbf{Curriculum scheduler.}
Following the most works~\cite{xu-etal-2020-curriculum, platanios-etal-2019-competence}, we schedule the curriculum at a linear pace (every 1,000 training iterations).
During each curriculum schedule, we enlarge the training set two times until it includes all the samples.
The capability upper bound $\overline{c_i^{s+1}}$ for $s+1$ stage by exponential growth: $\overline{c_i^{s+1}} = max\{ \gamma c_i^s, 1.0 \}$.
After that, we use criterion $v_i(x) < \overline{c_i^{s+1}}$ to construct candidate set $D_i^{s+1}$ for the $i$-th capability on the state $s+1$,
and use absolute contribution of $v_i$ to $F_{logits}$ as sampling ratio (\ie $\rho_1 : \rho_2 : \rho_3 : \rho_4$) to construct $D^{s+1}$.

\section{Experiments}
\label{sec:experiments}

\noindent
\textbf{Datasets.} \
We employ two question styles to evaluate our CBBC: answer span extraction and multiple choice. The former consists of SQuADv1~\cite{rajpurkar2016squad}, SQuADv2~\cite{rajpurkar2018know} and HotpotQA~\cite{yang2018hotpotqa}, while the latter adopts RACE~\cite{lai2017race}. For each dataset, we train and evaluate the model on official training and dev split, respectively.

%

\noindent
\textbf{Implementation details.} \
The source code and hyperparameters are included in the supplementary material.
We use BERT-base~\cite{devlin_bert_2018} as our backbone model, which is initialized by pre-trained parameters from cased BERT.
AdamW~\cite{loshchilov2017decoupled} optimizer with weight decay $5e-4$ and epsilon 8 is used to finetune the model with max sequence length $384$, document stride $128$.
The learning rate warms up over the first $10\%$ steps and then decays linearly to $0$ for all experiments with training batch size $16$ and maximum iteration $40,000$.

\noindent
\textbf{Baseline models.} \
In addition to the BERT-base model, we also consider the following ten baselines.
The first two baselines are trained through a pre-defined curriculum learning strategy, which sorts the samples, then feeds them to the model stage-by-stage.
``B+CL+$\mathcal{V}$ ($M_2$)'' sorts the samples by four capability-specific scores in an easy-to-difficult order.
``B+antiCL+$\mathcal{V}$ ($M_3$)'' does like ``$M_2$'', but in a reverse difficult-to-easy order.
The following five baselines are trained using our CBBC strategy to maximize the data value in each dimension, respectively.
``B+C+$v_1$ ($M_4$)'',
``B+C+$v_2$ ($M_5$)'',
``B+C+$v_3$ ($M_6$)''
and ``B+C+$v_4$ ($M_7$)'' use the corresponding $v_1$, $v_2$, $v_3$ and $v_4$ respectively to perform the competency test and filter samples.
``B+C+$\mathcal{V}_{corr}$ ($M_8$)'' is trained using four correlated scores through CBBC.
The following three baselines are devised by embedding other instance scoring methods into our CBBC pipeline.
``B+C+DatasetMap ($M_9$)'',
``B+C+Forgetting ($M_{10}$)''
and ``B+C+Predictability ($M_{11}$)'' substitute the capability-specific scores with the confidence score of true answer span~\cite{swayamdipta-etal-2020-dataset}, number of ``forgotten'' events~\cite{toneva2018empirical} and predictability score~\cite{le2020adversarial}, respectively.
The last two baselines (denoted as $M_{12}$ and $M_{13}$) are start-of-the-art curriculum learning pipelines consisting of DRCA~\cite{xu-etal-2020-curriculum} and CBCL~\cite{platanios-etal-2019-competence}.
Finally, our full model is trained using four decorrelated scores through CBBC instead. The critical difference between the full model and $M_8$ is the decorrelation operation.

\begin{table}[!t]
    \centering
     \resizebox{1.0\columnwidth}{!}{
        \begin{tabular}{c|c|cc|cc|cc|c}
            \toprule
            \multirow{2}{*}{\textbf{Name}}& \multirow{2}{*}{\textbf{Method}}& \multicolumn{2}{c|}{\textbf{SQuADv1}}& \multicolumn{2}{c|}{\textbf{SQuADv2}}& \multicolumn{2}{c|}{\textbf{HotpotQA}}& \textbf{RACE} \\
            & & \textbf{$EM$}& \textbf{$F_1$}& \textbf{$EM$}& \textbf{$F_1$}& \textbf{$EM$}& \textbf{$F_1$}& \textbf{$Acc.$} \\
            \midrule
            \midrule
            $M_1$& B& 81.25& 88.41& 77.32& 80.31& 63.53& 76.42& 63.02 \\
            \hline
            $M_2$& B+CL+$\mathcal{V}$& 84.21& 90.12& 81.16& 84.28& 66.44& 78.71& 66.68 \\
            $M_3$& B+antiCL+$\mathcal{V}$& 79.80& 87.31& 75.43& 79.19& 63.01& 75.03& 61.80 \\
            \hline
            $M_4$& B+C+$v_1$& 82.47& 89.29& 78.56& 81.62& 64.18& 76.88& 63.92 \\
            $M_5$& B+C+$v_2$& 82.67& 89.30& 78.59& 81.98& 64.12& 77.01& 64.11 \\
            $M_6$& B+C+$v_3$& 83.66& 89.68& 80.47& 83.23& 65.36& 78.75& 65.82 \\
            $M_7$& B+C+$v_4$& 85.03& 89.90& 81.60& 84.21& 66.30& 79.13& 65.73 \\
            $M_8$& B+C+$\mathcal{V}_{corr}$& 87.25& 91.65& 84.98& 87.64& 68.39& 80.42& 68.66 \\
            \hline
            $M_9$& B+C+DatasetMap& 82.04& 89.05& 78.04& 81.11& 63.86& 76.57& 63.51 \\
            $M_{10}$& B+C+Forgetting& 83.51& 89.77& 79.62& 82.99& 64.75& 77.62& 64.91 \\
            $M_{11}$& B+C+Predictability& 84.65& 90.67& 81.47& 84.22& 66.36& 79.75& 66.82 \\
            \hline
            $M_{12}$& B+DRCA& 85.05& 90.59& 82.19& 85.30& 67.07& 79.32& 67.48 \\
            $M_{13}$& B+CBCL& 86.15& 90.89& 83.17& 86.85& 67.72& 79.65& 67.79 \\
            \hline
            & \textbf{Ours}& \textbf{89.71}& \textbf{93.18}& \textbf{87.64}& \textbf{90.51}& \textbf{69.82}& \textbf{82.57}& \textbf{71.01} \\
            \bottomrule
        \end{tabular}
     }
    \caption{
    Quantitative results on four benchmark datasets.
    B and C represent the BERT backbone and our CBBC strategy, respectively.
    The best results are highlighted in \textbf{bold}.
    }
    \vspace{-5mm}
    \label{tab:comparisons}
\end{table}

\subsection{Experimental Results}
\noindent
\textbf{Quantitative Results.} \
We present a summary of our quantitative results in Table~\ref{tab:comparisons}.
As shown in the table, we have the following key observations.

On the one hand, our proposed competency framework does benefit the MRC learning efficiency in either a single or multiple dimensions.
For example, when using a pre-defined curriculum strategy, $M_2$ achieves $EM$ and $F_1$ far beyond $M_1$, highlighting that our quantification to data properties properly estimates the learning value contained in the data.
$M_3$ degrades performance w.r.t. $M_1$, demonstrating that the learning strategy from easy to difficult samples is more reasonable than the reverse.
When equipped with our CBBC, all models of $M_4$, $M_5$, $M_6$ and $M_7$ achieve improvements w.r.t. $M_1$ on four datasets, which indicates the significant contribution of each capability dimension on gradually increasing the model capability.
In particular, among the four different dimensions, $M_7$ has the best result, indicating that understanding sentences is a relatively more important capability for MRC.
$M_8$ outperforms all the models except for ours. This demonstrates that our CBBC can maximize the learning value of the data sample to increase an MRC model's capability.

On the other hand, our framework wins other scoring methods and curriculum learning pipelines by a considerable margin.
Although $M_9$, $M_{10}$, $M_{11}$, $M_{12}$ and $M_{13}$ achieve substantial improvements on four datasets w.r.t. $M_1$, their performances are still worse than our full model.
These results verify that our proposed framework can assess the model capability more correctly and make better use of the learning value within data.

Finally, our full model achieves significantly higher $EM$, $F_1$ and $Acc.$ compared to all other baselines, demonstrating the necessity of the decorrelation between capability-specific scores.
Its superior performance roots from constructing a decorrelated value representation of each dimension for our CBBC learning strategy.
Overall, compared to $M_1$, our full model achieves tremendous improvement of $EM$ / $F_1$ up to 11.22\% / 8.71\% on the average of three answer extraction style datasets.


\begin{figure}[!t]
\centering
	\begin{subfigure}{0.49\columnwidth}
		\includegraphics[width=\textwidth]{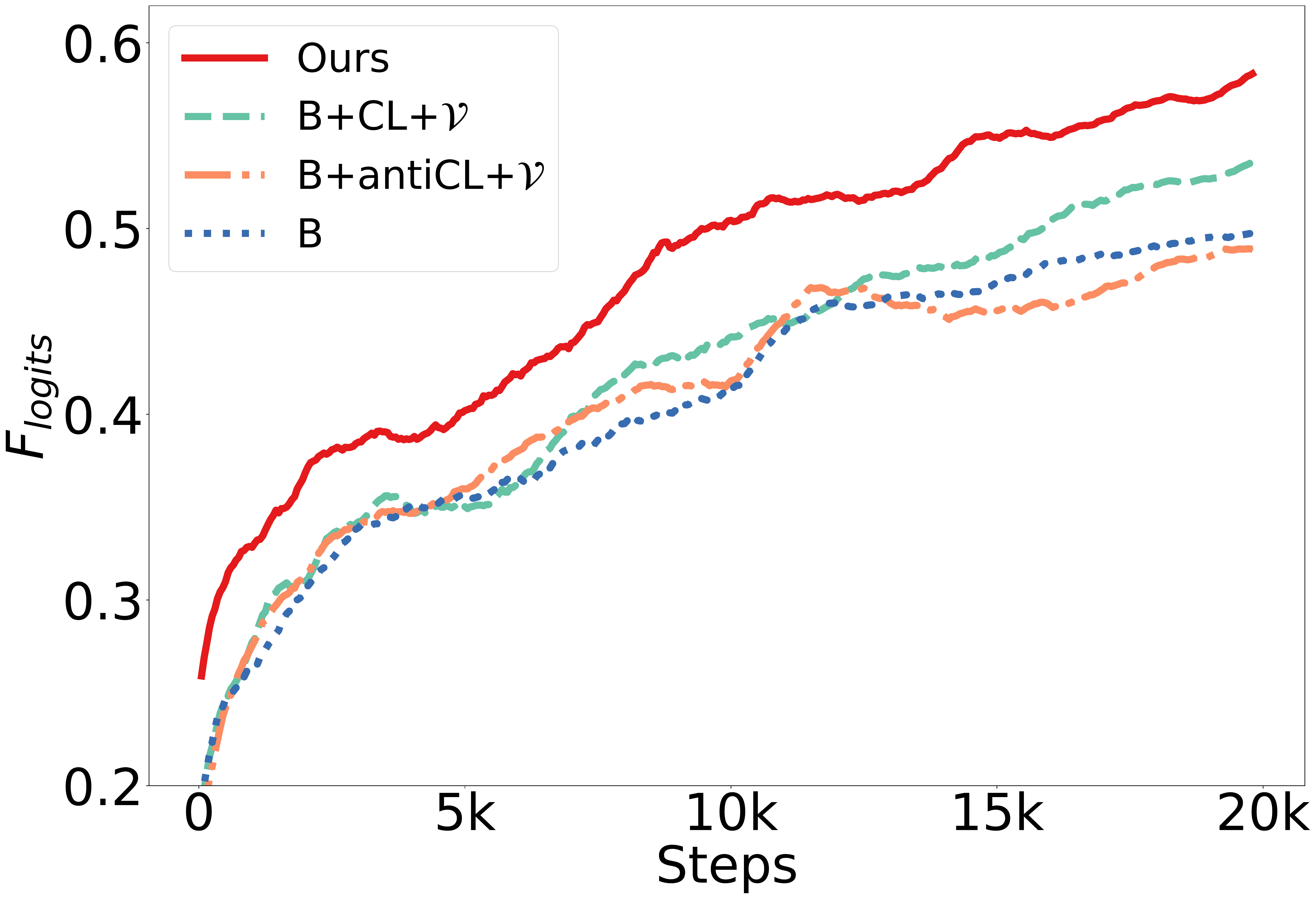}
	\end{subfigure}
    \hfill
	\begin{subfigure}{0.49\columnwidth}
		\includegraphics[width=\textwidth]{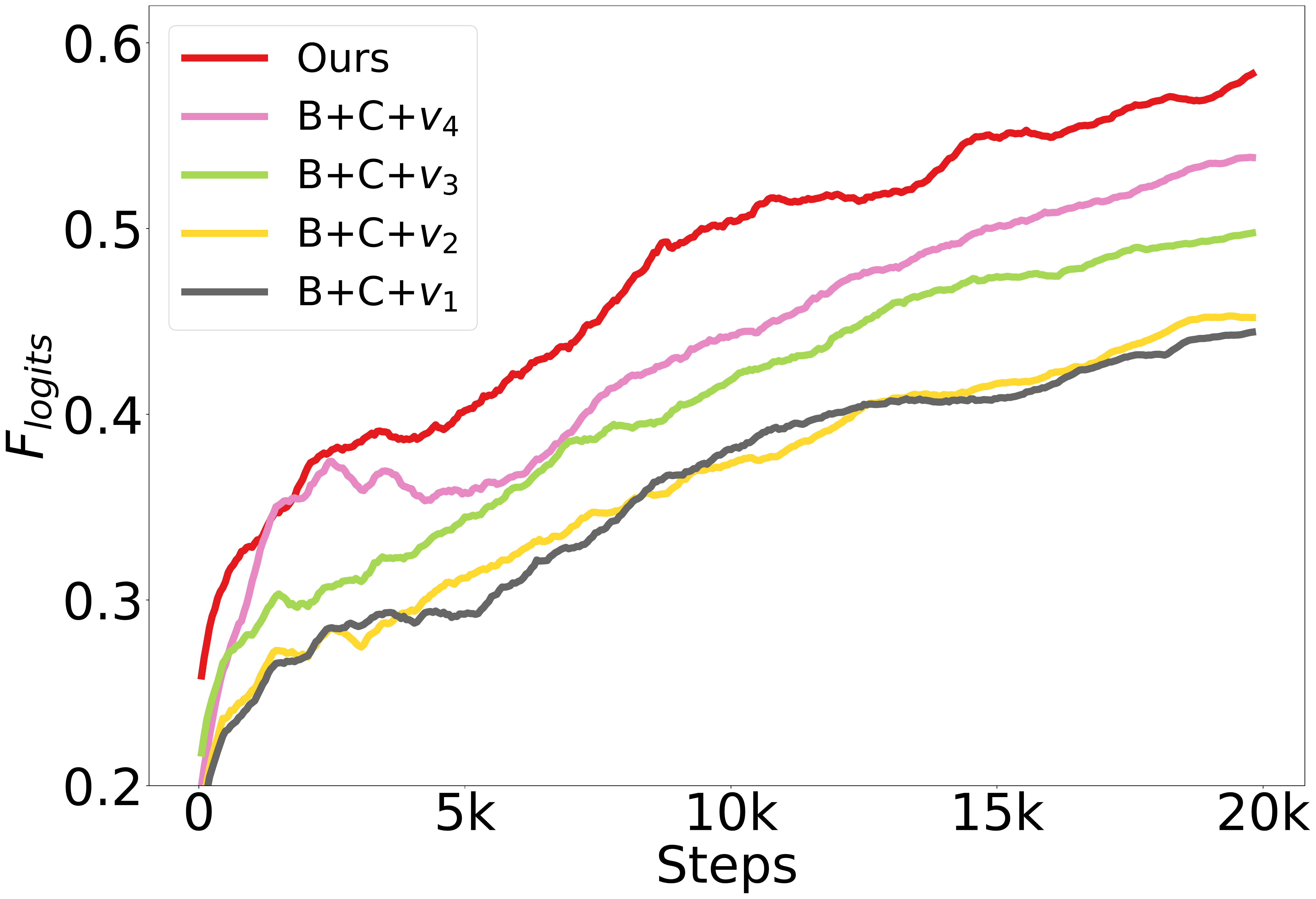}
	\end{subfigure}

    \caption{Illustration of performance (smoothed by averaging $F_{logits}$ every 32 steps) of various baseline models on HotpotQA dev split as training progresses.
    Ours (denoted as \textcolor{dred}{red} plot) outperforms the other models from the start of training.}
    \vspace{-2mm}
    \label{fig:training-fl}
\end{figure}

\begin{figure}[t]
	\centering

	\begin{subfigure}{0.488\columnwidth}
    \includegraphics[width=\textwidth]{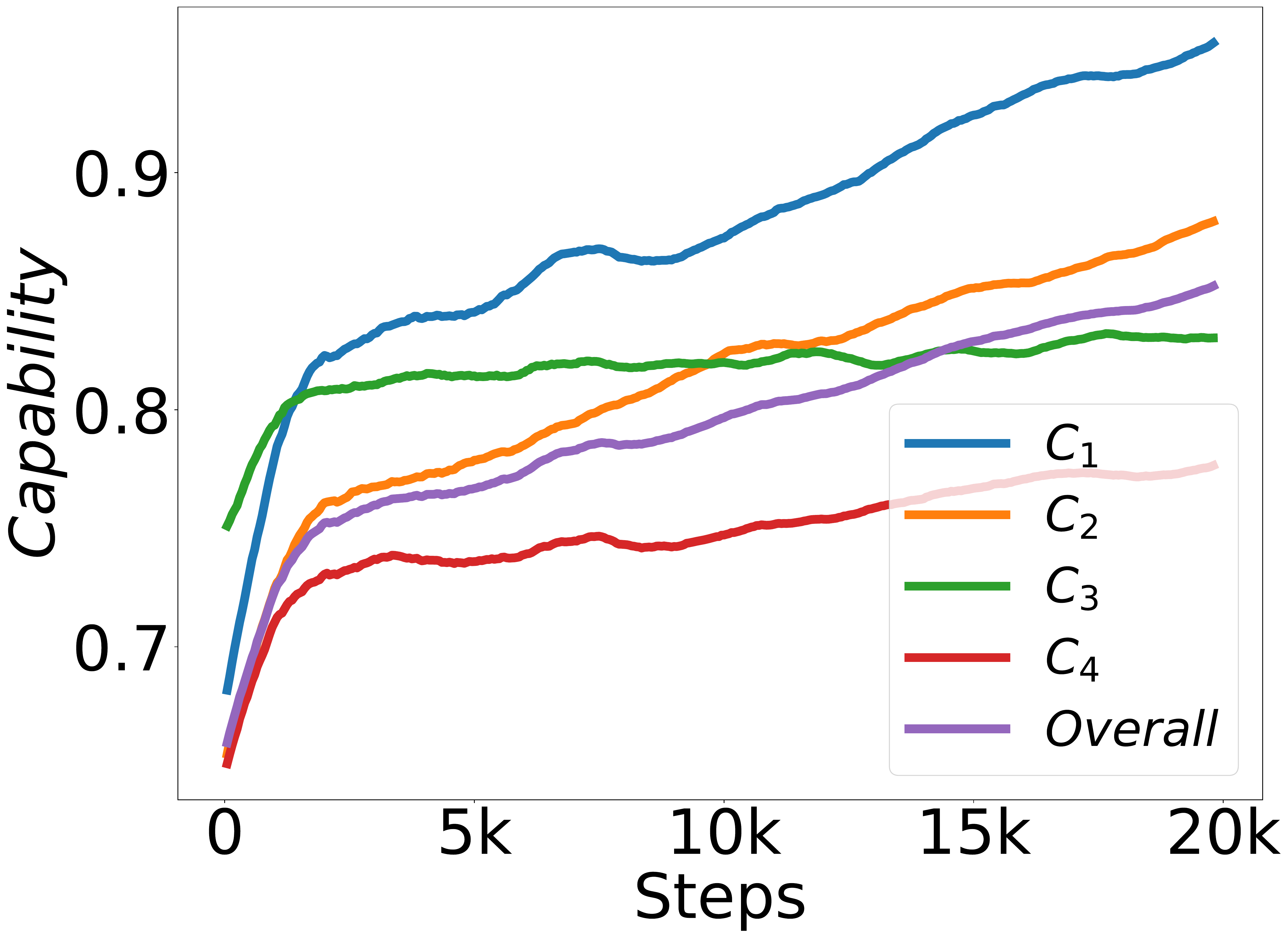}
    \subcaption{Plots illustrating changes of each MRC capability during training.}
	\end{subfigure}
	\hfill
	\begin{subfigure}{0.47\columnwidth}
    \includegraphics[width=\textwidth]{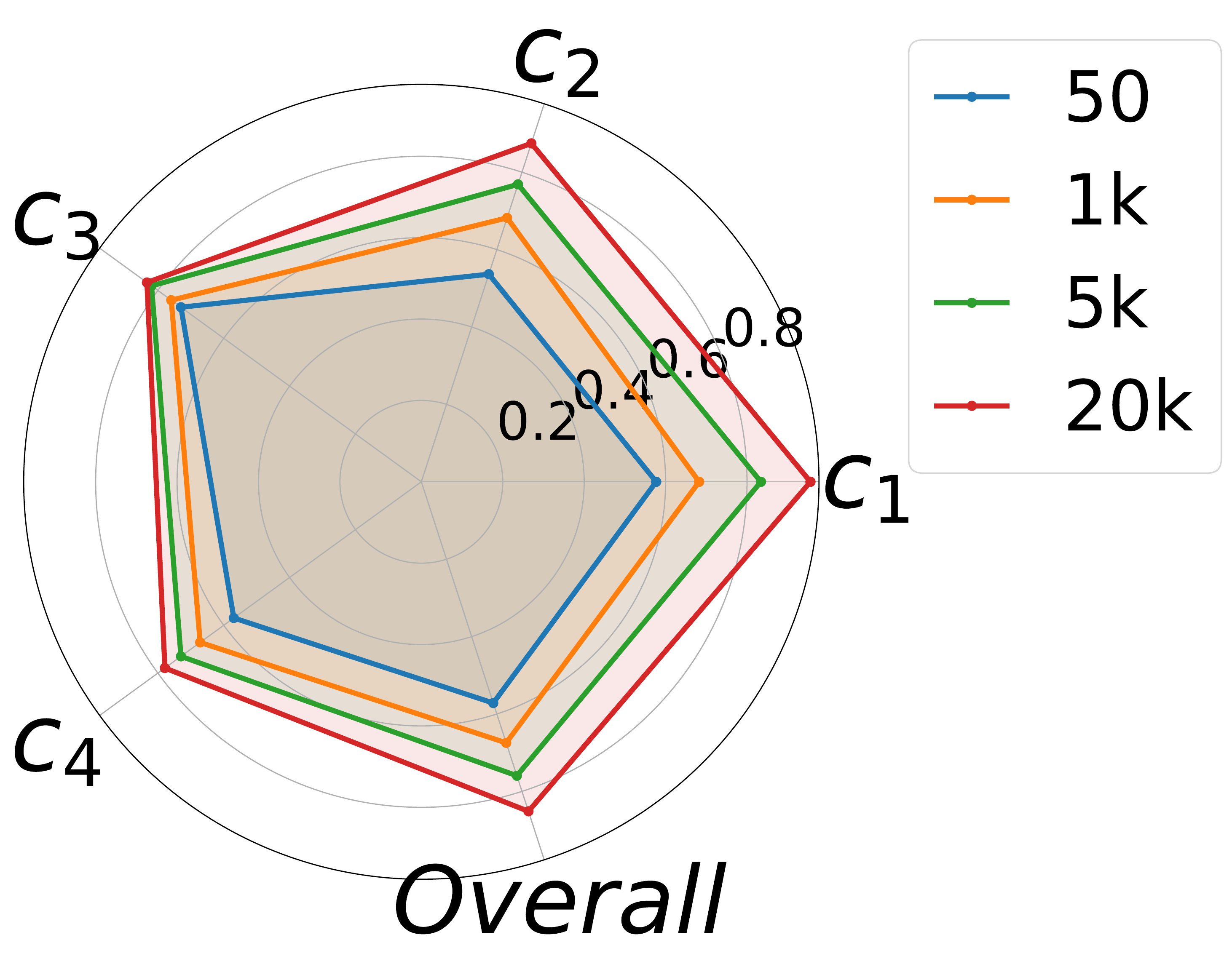}
    \subcaption{A 5-dimensional map of MRC model capabilities on step 50, 1k, 2k, 20k.}
	\end{subfigure}

    \caption{ Illustration of MRC model capabilities on different training stages.}
     \label{fig:capability-map}
    \vspace{-5mm}
\end{figure}

\noindent
\textbf{Qualitative Results.} \
Figure~\ref{fig:training-fl} shows the performance of baselines on the HotpotQA dev set.
There are two observations worth noting here.
First, the performance of our full model lies consistently on top of the other baseline models during the whole training stage. This result shows that CBBC can make the model more prepared for complex samples by enlarging its capability boundary step by step.
Second, the performance plot of the baseline model with $v_4$ sits on top of other baselines with $v_1$, $v_2$, and $v_3$ from the beginning of training to the end. This result highlights the main contribution of $v_4$ (understanding sentences) to the final performance.

Furthermore, the capability map after max-min normalization of the model is shown in Figure~\ref{fig:capability-map}.
First, among 4-dimensional capability, the $c_3$ (\ie understanding words) has the largest initial value.
A possible explanation is that pre-trained BERT has a fair amount of prior knowledge obtained from unlabeled corpus, which concentrates more on semantic understanding of words.
Second, the capability $c_1$ increases at the fastest speed as the training progresses. Interestingly, the model $M_4$ based on $v_1$ does not seem improving accordingly as the capability $c_1$ increases.
The possible reason could be that the superficial structure is easy to learn from samples but makes a limited contribution to the final performance.
\emph{Please refer to Appendix~\ref{sec:additional-models} for the results of other MRC models.}

\begin{table}[!t]
    \resizebox{1.0\columnwidth}{!}{
        \begin{tabular}{c|c|c|c|c}
            \toprule
            \large
            \textbf{Value}& \textbf{SQuADv1}& \textbf{SQuADv2}& \textbf{HotpotQA}& \textbf{RACE} \\
            \midrule
            
            \multirow{2}{*}{$v_1$}& 0.602& 0.536& 0.630& 0.589 \\
            & 0.625& 0.559& 0.653& 0.612 \\
            \hline
            
            \multirow{2}{*}{$v_2$}& 0.696& 0.683& 0.755& 0.620 \\
            & 0.714& 0.701& 0.773& 0.638 \\
            \hline
            
            \multirow{2}{*}{$v_3$}& 0.730& 0.573& 0.550& 0.797 \\
            & 0.674& 0.517& 0.494& 0.741 \\
            \hline
            
            \multirow{6}{*}{$v_4$}& 0.553& 0.674& 0.672& 0.819 \\
            & 0.488& 0.609& 0.607& 0.754 \\
            & 0.468& 0.589& 0.587& 0.734 \\
            & 0.472& 0.593& 0.591& 0.738 \\
            & 0.547& 0.668& 0.666& 0.813 \\
            & 0.503& 0.612& 0.609& 0.761 \\

            \bottomrule
        \end{tabular}
    }
    \caption{
    Pearson’s correlation coefficients between human judgments and data properties for four capabilities and their subclasses.
    }
    \vspace{-5mm}
    \label{tab:annotation-correlation}
\end{table}

\subsection{Human Annotation}

\noindent
\textbf{Annotation specification.} \
we ask three annotators to answer ($100 \times 4 = 400$) questions randomly sampled from four datasets, consisting of SQuADv1, SQuADv2, HotpotQA, and RACE.
Using only our proposed four capabilities, they first read the context, question, and gold standard answer (the correct candidate answer under multiple-choice situation), and then choose the evidence sentences in context.
After that, they respectively label the subclasses of four major capabilities as 1 (required) or 0 (not required).
\emph{Please refer to Appendix~\ref{sec:annotation-details} for more details about human annotation.}

\noindent
\textbf{Annotation results.} \
In the annotation of required capabilities, the inter-annotator agreement is 75.33\% for all 400 samples.
We use the average of three corresponding annotator labels as the final human judgments for a specific sub-capability required by the question.
Finally, a sample will be annotated ($2+2+2+6 = 12$) human ratings.
Table~\ref{tab:annotation-correlation} summarizes the correlations between human judgments and capability-specific scores of samples.
The relatively strong correlations on all four dimensions indicate that our employed heuristic metrics can reasonably approximate the learning value contained in the samples.

\section{Related Work}
\label{sec:related}

%

\noindent
\textbf{Analytic approaches to MRC capability.} \
Some works performed skill-based analyses for the MRC model.
In the scientific question domain, \citet{Clark2018ThinkYH} constituted the ARC benchmark, which requires far more powerful knowledge
and reasoning than previous benchmarks.
In a generalizable definition, \citet{sugawara2017prerequisite} proposed a set of 10 skills for MCTest~\cite{richardson2013mctest}. 
The others focused more on the analysis of the MRC dataset itself.
For example,
\citet{sugawara2020assessing} proposed a semi-automated, ablation-based methodology to assess the capacities of datasets.
\citet{rajpurkar2016squad} analyzed their proposed datasets using several types of reasoning, \eg lexical and syntactic variation, and multiple sentence reasoning.
Nevertheless, they require too costly human efforts and ignore that the model capability changes as training progresses. 

\noindent
\textbf{Data selection for debiased representations.} \
Some works proposed different criteria to score instances according to the model response to input.
\citet{swayamdipta-etal-2020-dataset} built data maps using training dynamics measures for scoring data samples.
\citet{toneva2018empirical} also employed the number of ``forgotten'' events to measure a sample, which was misclassified during a later epoch of training, despite being classified correctly earlier.
The others~\cite{le2020adversarial} used adversarial filtering algorithms to rank instances based on their ``predictability''. 
However, these approaches require training a model once in advance on the dataset to obtain the corresponding training dynamics, which is computationally expensive, especially when using a large model.


\section{Conclusion}
\label{sec:conclusion}
We design a competency assessment framework for MRC capabilities, which describes model skills in an explainable and multi-dimensional manner.
By leveraging the framework, we further uncover and disentangle the connections between various data properties and model performance on a specific task, as well as propose a capability boundary breakthrough curriculum (CBBC) strategy to maximize the data value and improve training efficiency.
The experiments performed on four benchmark datasets verified that our approach can significantly improve the performance of existing MRC models.
Our work shows a deep understanding of model capabilities and data properties helps monitor the model skills during training and improves learning efficiency.
Our framework and learning strategy are also generally applicable to other NLP tasks.

\section*{Acknowledgements}
This work has been supported in part by National Key Research and Development Program of China (2018AAA0101900), Zhejiang NSF (LR21F020004), Key Research and Development Program of Zhejiang Province, China (No. 2021C01013), Alibaba-Zhejiang University Joint Research Institute of Frontier Technologies, Chinese Knowledge Center of Engineering Science and Technology (CKCEST).


\bibliography{anthology,custom}
\bibliographystyle{acl_natbib}

\appendix

\section{Annotation Details}
\label{sec:annotation-details}
We ask three annotators to answer ($100 \times 4 = 400$) questions randomly sampled from four datasets, consisting of SQuADv1, SQuADv2, HotpotQA, and RACE.
They are graduate students majoring in Computer Science or Electronic Engineering and competent in English.
They voluntarily offer to help without being compensated in any form.
Before annotation, they are informed of the detailed annotation instruction with the following three steps.
\begin{itemize}
    \item \textbf{Step 1.} Make a hypothesis using a question statement and gold standard answer or the correct candidate answer under the multiple-choice situation.
    
    Example 1:
    
    Q: Why did Tom look angry? A: His sister ate his cake.
    
    $\rightarrow$ \emph{Hypothesis}: Tom looked angry because his sister ate his cake.
    
    Example 2:
    
    Q: When did French Revolution occur? A: In 1789
    
    $\rightarrow$ \emph{Hypothesis}: French Revolution occurred in 1789.
    
    \item \textbf{Step 2.} Select sentences (from the context) required to provide the hypothesis.
    
    Example 1:
    
    \emph{Context}: (C1) Tom is a student. (C2) Tom looks annoyed because his sister ate his cake. (C3) His sister's name is Sylvia.
    
    \emph{Hypothesis}: Tom looks angry because his sister ate his cake.
    
    $\rightarrow$ \emph{Select}: C2
    
    \item \textbf{Step 3.} Select capabilities required for understanding an entailment from selected context sentences to hypothesis and label the corresponding capability as 1 (required).
    
    Example 1:
    
    C2: Tom looks annoyed because his sister ate his cake.
    
    \emph{Hypothesis}: Tom looks angry because his sister ate his cake.
    
    $\rightarrow$ \emph{Capability}: causal relation (``because''), semantic overlap (lexical knowledge of ``annoyed = angry'')
\end{itemize}
Then, we describe our annotation schema in greater detail.
We present the respective phenomenon, give a short description, and present an example illustrating the corresponding feature.
\begin{itemize}
\item
\textbf{Reading words}

\noindent
\textbf{Recognize vocabulary.} We annotate this as ``1'' if repetition of some word rarely occurs in a sentence (less than two times in every ten words).

\emph{Question with label 0}: The creek of which Paradise Creek is a tributary of what river?

\emph{Context}: Paradise Creek is a 9.6 mi tributary of Brodhead Creek in the Poconos of eastern Pennsylvania in the United States. Brodhead Creek is a 21.9 mi tributary of the Delaware River in the Poconos of eastern Pennsylvania in the United States.

\emph{Question with label 1}: Of these two publications--Báiki and Sick--what type of publication is the one that was published most frequently?

\emph{Context}: Báiki: The International Sámi Journal ("Báiki" means a place in Sami) is a biannual English-language publication that covers Sami culture, history, and current affairs. The coverage also includes the community affairs of the Sami in North America, estimated at some 30,000 people. Sick was a satirical-humor magazine published from 1960 to 1980, lasting 134 issues.

\noindent
\textbf{Recognize function words.} We annotate this as ``1'' if a sentence consists of lots of the structural relationships between words signaled by function words (more than five times in every ten words).

\emph{Question}: What drug is among the list of illegal drugs in the Philippines and can be taken by mouth or by injection?

\emph{Context}: $[\ldots]$ Ephedrine and methylenedioxy methamphetamine are also among the list of illegal drugs that are of great concern to the authorities. Ephedrine is a medication and stimulant. $[\ldots]$

\item
\textbf{Reading sentences}

\noindent
\textbf{Recognize grammaticality.} We annotate this as ``1'' the sentence pattern and grammar involved are relatively complex, such as multiple nested subordinate clause structures.

\emph{Question}: Sudha Kheterpal, who is a musician best known as the percussionist in Faithless, has played with what singer who is recognized as the highest-selling Australian artist of all time by the Australian Recording Industry Association?

\noindent
\textbf{Readability.} We annotate this as``1'' if there are a lot of obscure words in the question or context (more than five times in every ten words).

\emph{Question}: The creature HNoMS Draug is named after comes from what kind of mythology?

\emph{Context}: Two ships of the Royal Norwegian Navy have borne the name HNoMS ``Draug'', after the sea revenant Draugr: The draugr or draug (Old Norse: ``draugr'', plural draugar ; modern Icelandic: ``draugur'', Faroese: ``dreygur'' and Danish, Swedish, and Norwegian: ``draug'' ), also called aptrganga or aptrgangr , literally ``again-walker'' (Icelandic: ``afturganga'' ) is an undead creature from Norse mythology.

\item
\textbf{Understanding words}

\noindent
\textbf{Arithmetic operation.} We annotate this as ``1'' if an arithmetic operation needs to be performed to answer the question, such as addition, subtraction, ordering, and counting.

\emph{Question}: How many points were the Giants behind the Dolphins at the start of the 4th quarter?

\emph{Context}: New York was down 17-10 behind two rushing touchdowns.

\noindent
\textbf{Logical operation.} We annotate this as ``1'' if it is required to understand the concept of quantification (existential and universal) in order to determine the correct answer.

\emph{Question}: How many presents did Susan receive?

\emph{Context}: On the day of the party, all five friends showed up. Each friend Quantification had a present for Susan.

\item
\textbf{Understanding sentences}

\noindent
\textbf{Syntactic and semantic overlap.} We annotate this as ``1'' if some part of the context and the question overlap semantically.

\emph{Question}: Is it freezing today?

\emph{Context}: The weather is cold today.

\noindent
\textbf{Coreference resolution.} We annotate this as ``1'' if inter-sentence coreference and anaphora need to be resolved in order to retrieve the expected answer.

\emph{Question}: What is the name of the psychologist who is known as the originator of social learning theory?

\emph{Context}: Albert Bandura OC (born December 4, 1925) is a psychologist who is the David Starr Jordan Professor Emeritus of Social Science in Psychology at Stanford University. $[\ldots]$ He is known as the originator of social learning theory and the theoretical construct of self-efficacy and is also responsible for the influential 1961 Bobo doll experiment.

\noindent
\textbf{Con/Dis-junction, negation.} We annotate this as ``1'' if logical conjunction, disjunction, or negation needs to be resolved in order to conclude the answer.

\emph{Question}: Is dad in the living room?

\emph{Context}: Dad is either in the kitchen or in the living room.

\emph{Question}: How many percent are not Marriage couples living together?

\emph{Context}: $[\ldots]$ 46.28\% were Marriage living together. $[\ldots]$

\noindent
\textbf{Causality.} We annotate this as ``1'' if causal (\ie cause-effect) reasoning between events, entities, or concepts is required to answer a question correctly.

\emph{Question}: Why did Sam stop Mom from making four sandwiches?

\emph{Context}: $[\ldots]$ There are three of us, so we need three sandwiches. $[\ldots]$

\noindent
\textbf{Spatial/Temporal relationship.}  We annotate this as ``1'' understanding about directions, environment, spatiality, and succession is required in order to derive an answer.

\emph{Question}: What is the 2010 population of the city 2.1 miles southwest of Marietta Air Force Station?

\emph{Context}: Marietta Air Force Station is located 2.1 mi northeast of Smyrna, Georgia.

\noindent
\textbf{Multi-hop reasoning.} We annotate this as ``1'' if information to answer the question needs to be gathered from multiple supporting facts, ``Multi-hop'' by commonly mentioned entities, concepts, or events.
This phenomenon is also known as ``Bridging'' in literature.

\emph{Question}: What show does the host of The 2011 Teen Choice Awards ceremony currently star on?

\emph{Context}: $[\ldots]$ The 2011 Teen Choice Awards ceremony, hosted by Kaley Cuoco, aired live on August 7, 2011, at 8/7c on Fox. $[\ldots]$ Kaley Christine Cuoco is an American actress. Since 2007, she has starred as Penny on the CBS sitcom ``The Big Bang Theory'', for which she has received Satellite, Critics’ Choice, and People’s Choice Awards. $[\ldots]$
\end{itemize}

\begin{table*}[!t]
    \resizebox{1.0\textwidth}{!}{
        \begin{tabular}{c|c|c}
            \toprule
            
            \textbf{ID}& Q1(5a7322a25542991f9a20c634)& Q2(5a72bd0b5542992359bc318f) \\
            
            \hline
            
            \multirow{22}{*}{\rotatebox{90}{\textbf{Context}}}& \multirow{22}{7cm}{\centering \textcolor{dgreen}{The Metropolitan Life Insurance Company Tower}, colloquially known as the Met Life Tower, is a landmark skyscraper located on Madison Avenue near the intersection with East 23rd Street, across from Madison Square Park in Manhattan, New York City. Designed by the architectural firm of Napoleon LeBrun \& Sons and built by the Hedden Construction Company, the tower is modeled after the Campanile in Venice, Italy. The hotel located in the clock tower portion of the building has the address 5 Madison Avenue, while the office building covering the rest of the block, occupied primarily by Credit Suisse, is referred to as 1 Madison Avenue. 15 Hudson Yards is a residential building currently under construction on Manhattan's West Side. Located in Chelsea near Hell's Kitchen Penn Station area, the building is a part of the Hudson Yards project, a plan to redevelop the Metropolitan Transportation Authority's West Side Yards. The tower started construction on December 4, 2014.}&
            \multirow{22}{7cm}{\centering Andrea Louise Martin (born January 15, 1947) is an American actress, singer, author, and comedian, best known for her work in the television series "SCTV". She has appeared in films such as "Black Christmas" (1974), "Wag the Dog" (1997), "Hedwig and the Angry Inch" (2001), "My Big Fat Greek Wedding" (2002), and "My Big Fat Greek Wedding 2" (2016), and lent her voice to the animated films "Anastasia" (1997), "The Rugrats Movie" (1998) and " (2001).\textcolor{dgreen}{ Mark S. Hoplamazian} is an American businessman who is the President and CEO of Hyatt Hotels Corporation. He received his A.B. in economics from Harvard College and his M.B.A. from the University of Chicago Booth School of Business.} \\
            & &\\ & &\\ & &\\ & &\\ & &\\ & &\\ & &\\ & &\\ & &\\ & &\\ & &\\ & &\\ & &\\ & &\\ & &\\ & &\\ & &\\ & &\\ & &\\ & &\\ & &\\ & &\\
            
            \hline

            \multirow{4}{*}{\rotatebox{90}{\textbf{Question}}}& \multirow{4}{7cm}{\centering Was the Metropolitan Life Insurance Company Tower [Met Life Tower] or the 15 Hudson Yards building designed by the firm of Napoleon LeBrun \& Sons?}&
            \multirow{4}{7cm}{\centering Who achieved more academically, Andrea Martin or Mark Hoplamazian?} \\
            & & \\ & & \\ & & \\
            
            \bottomrule
        \end{tabular}
    }
    \caption{Two samples from HotpotQA dev set.
    The answer span in the context is marked in \textcolor{dgreen}{green}.}
    \label{tab:examples}
\end{table*}

\begin{table*}[!t]
    \resizebox{1.0\textwidth}{!}{
        \begin{tabular}{c|c|cc|cc}
            \toprule
            
            \multirow{2}{*}{\textbf{Value}}& \multirow{2}{*}{\textbf{Metrics}}& \multicolumn{2}{c|}{\textbf{Q1(5a7322a25542991f9a20c634)}}& \multicolumn{2}{c}{\textbf{Q2(5a72bd0b5542992359bc318f)}} \\
            &  & raw& normalized& raw& normalized \\
            
            \hline
            
            \multirow{3}{*}{$v_1$}& intra1& 0.521& 0.252& 0.608& \textbf{0.681} \\
            & entropy1& 6.243& 0.530& 6.363& \textbf{0.636} \\
            & ntopwrods& 0.268& 0.509& 0.274& \textbf{0.575} \\
            
            \hline
            
            \multirow{3}{*}{$v_2$}& height& 9.000& \textbf{0.904}& 7.000& 0.809 \\
            & flesch kincaid& 16.748& \textbf{0.913}& 10.574& 0.196 \\
            & ari& 17.659& \textbf{0.895}& 9.835& 0.119 \\
            
            \hline
            
            \multirow{2}{*}{$v_3$}& nnums& 0.039& 0.722& 0.051& \textbf{0.861} \\
            & nlogicals& 0.002& 0.506& 0.009& \textbf{0.735} \\
            
            \hline
            
            \multirow{7}{*}{$v_4$}& BERTScore& 0.862& 0.043& 0.698& \textbf{0.959} \\
            & MoverScore& 0.154& 0.022& -0.210& \textbf{0.900} \\
            & ncoreferences& 0.005& 0.079& 0.071& \textbf{0.991} \\
            & njunctions& 0.012& 0.069& 0.071& \textbf{0.957} \\
            & ncausals& 0.001& 0.275& 0.009& \textbf{0.619} \\
            & nspatialtemporals& 0.078& \textbf{0.779}& 0.035& 0.512 \\
            & nfacts& 2.000& 0.674& 3.000& \textbf{0.913} \\
            \bottomrule
        \end{tabular}
    }
    \caption{Two samples with our raw and normalized metrics. The higher normalized scores are marked in \textbf{bold}.}
    \vspace{-5mm}
    \label{tab:examples-metrics}
\end{table*}

\section{Examples of Our Employed Metrics}
\label{sec:examples-metrics}
We first present a brief overview of employed metrics, especially those that are adapted from other studies.
In the following descriptions, the question $Q$ and corresponding context $C$ are denoted as the sequence of n-grams $Q^n = \{ q^n_i \}$ and $C^n = \{ c^n_j \}$, respectively.

\noindent
\textbf{Intra-n and Ent-n.} \emph{Intra-n}~\cite{gu2018dialogwae} and \emph{Ent-n}~\cite{serban2017hierarchical} are originally designed to evaluate the diversity of neural dialogue responses.
The former calculates the ratio of distinct unigrams (Intra-1) and bigrams (Intra-2) in generated responses, while the latter measures the information entropy of n-grams.
Specifically, in our work, they are formulated as:
\begin{align}
    \text{Intra-n} &= \frac{Unique(Q^n)}{\vert Q^n \vert} \\
    \text{Ent-n} &= \sum_{i=1}^{}{- \frac{Count(q^n_i)}{\vert Q^n \vert} \log \frac{Count(q^n_i)}{\vert Q^n \vert}}
\end{align}

\noindent
\textbf{Tree statistics.}
Empirically, most of the complicated sentences have a relatively high and wide constituency parsing tree.
In this paper, we calculate the height and width of the constituency parsing tree by Standford CoreNLP API~\cite{manning2014stanford}.

\noindent
\textbf{Readability metrics.}
Readability is the ease with which a reader can understand a written text.
Readability metrics produce an approximate representation of the US grade level needed to comprehend the text and are widely used in the field of education to assess the English proficiency of non-native English speakers.
We employ the py-readability package (\url{https://pypi.org/project/py-readability-metrics/}) to calculate the readability of a question based on two metrics, including \emph{Flesch Kincaid Grade Level} and \emph{Automated Readability Index (ARI)}.

\noindent
\textbf{BERTScore and MoverScore.}
To measure the semantic overlap between the question and corresponding context, we slightly modify the contextualized embedding-based similarity metrics of text generation task, comprising \emph{BERTScore}~\cite{zhang2020bertscore} and \emph{MoverScore}~\cite{zhao2019moverscore}.
Unlike the original implementation, the question and the context rather than the gold standard reference are fed into the metrics computation.

We exemplify two samples from HotpotQA dev set to show the difference of specific metrics in Table~\ref{tab:examples} and Table~\ref{tab:examples-metrics}.

\begin{figure}[!t]
	\centering
	
    \includegraphics[width=\columnwidth]{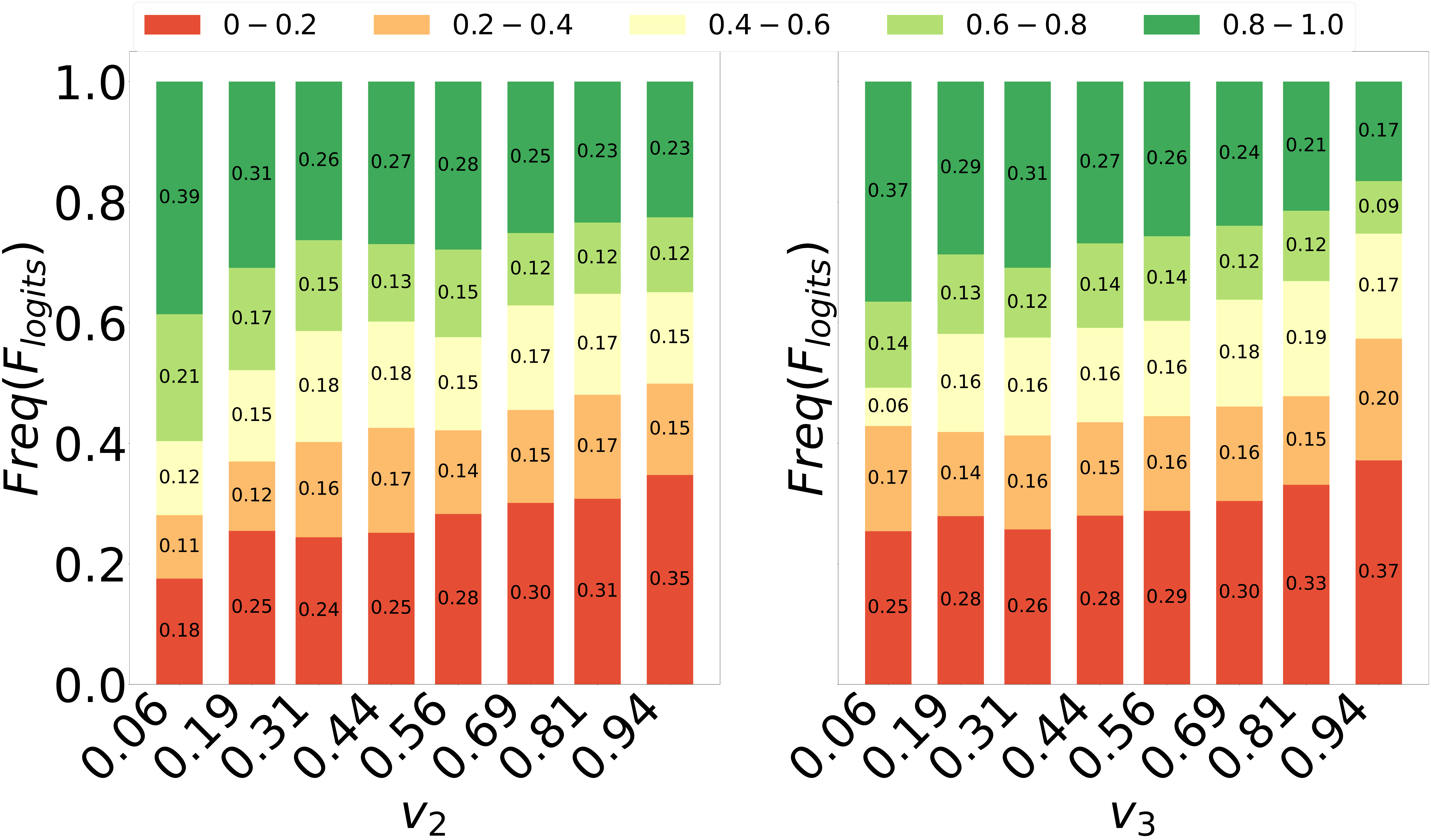}

	\caption{Bar diagram illustrating the relationship between the distribution of model performance and different ranges of $v_i$.
	Horizontal axes represent the different score ranges of $v_i$ of samples, and the vertical axis means the performance distribution by the frequency of $F_{logits}$ on five levels (plotted in five colors).
	}
	\vspace{-1mm}
	\label{fig:v23-value-distribution}
\end{figure}

\begin{figure}[!t]
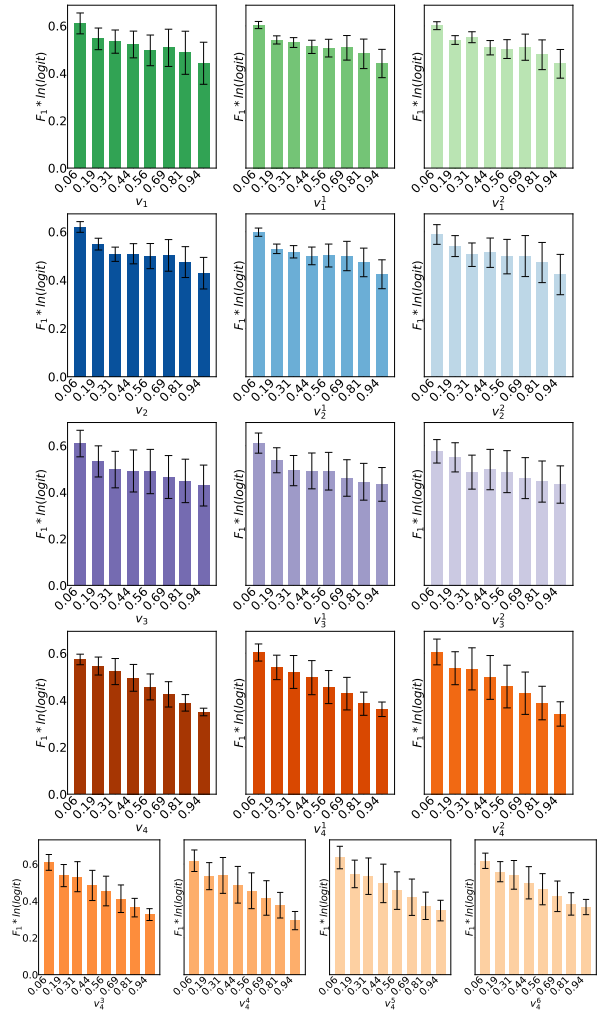

	\centering
	
	\begin{subfigure}{0.92\columnwidth}
		\includegraphics[width=\textwidth]{figures/fl-v1-12-mean.pdf}
	\end{subfigure}
    \\
	\begin{subfigure}{0.92\columnwidth}
		\includegraphics[width=\textwidth]{figures/fl-v2-12-mean.pdf}
	\end{subfigure}
    \\
	\begin{subfigure}{0.92\columnwidth}
		\includegraphics[width=\textwidth]{figures/fl-v3-12-mean.pdf}
	\end{subfigure}
	\\
	\begin{subfigure}{0.92\columnwidth}
		\includegraphics[width=\textwidth]{figures/fl-v4-12-mean.pdf}
	\end{subfigure}
	\\
	\begin{subfigure}{\columnwidth}
		\includegraphics[width=\textwidth]{figures/fl-v4-3456-mean.pdf}
	\end{subfigure}

	\caption{Bar diagram illustrating the relationship between the mean value and standard deviation of model performance and different ranges of $v_i$ and its subclasses.
	The height of each bar and its error line represent the mean value and standard deviation of model performance, respectively.
	}
	\label{fig:metrics-mean}
\end{figure}

\begin{table*}[!t]
    \centering
     \resizebox{0.7\textwidth}{!}{
        \begin{tabular}{c|cc|cc|cc|c}
            \toprule
            \multirow{2}{*}{\textbf{Method}}& \multicolumn{2}{c|}{\textbf{SQuADv1}}& \multicolumn{2}{c|}{\textbf{SQuADv2}}& \multicolumn{2}{c|}{\textbf{HotpotQA}}& \textbf{RACE} \\
            & \textbf{$EM$}& \textbf{$F_1$}& \textbf{$EM$}& \textbf{$F_1$}& \textbf{$EM$}& \textbf{$F_1$}& \textbf{$Acc.$} \\
            \midrule
            \midrule
            R-Net& 76.59& 85.83& 71.87& 75.10& 60.78& 73.17& 59.49 \\
            \hline
            R-Net+CBBC+$v_1$& 78.11& 86.36& 73.37& 77.15& 61.75& 73.80& 60.20 \\
            R-Net+CBBC+$v_2$& 78.23& 86.91& 73.40& 76.52& 61.03& 73.80& 59.92 \\
            R-Net+CBBC+$v_3$& 79.71& 87.63& 74.98& 78.40& 61.92& 74.86& 61.32 \\
            R-Net+CBBC+$v_4$& 80.27& 87.77& 76.35& 79.15& 62.85& 76.29& 62.63 \\
            R-Net+CBBC+$\mathcal{V}_{corr}$& 82.49& 88.47& 78.50& 81.15& 64.41& 77.28& 63.33 \\
            \hline
            \textbf{R-Net+CBBC+$\mathcal{V}$}& \textbf{83.87}& \textbf{89.75}& \textbf{80.85}& \textbf{83.56}& \textbf{65.87}& \textbf{77.96}& \textbf{65.47} \\
            \bottomrule
        \end{tabular}
     }
    \caption{Quantitative results of R-Net backbone on four benchmark datasets. The best results are highlighted in \textbf{bold}.}
    \vspace{-4mm}
    \label{tab:comparisons-r-net}
\end{table*}

\begin{table*}[!t]
    \centering
     \resizebox{0.7\textwidth}{!}{
        \begin{tabular}{c|cc|cc|cc|c}
            \toprule
            \multirow{2}{*}{\textbf{Method}}& \multicolumn{2}{c|}{\textbf{SQuADv1}}& \multicolumn{2}{c|}{\textbf{SQuADv2}}& \multicolumn{2}{c|}{\textbf{HotpotQA}}& \textbf{RACE} \\
            & \textbf{$EM$}& \textbf{$F_1$}& \textbf{$EM$}& \textbf{$F_1$}& \textbf{$EM$}& \textbf{$F_1$}& \textbf{$Acc.$} \\
            \midrule
            \midrule
            QANet& 77.85& 86.54& 73.42& 76.62& 61.72& 74.09& 60.69 \\
            \hline
            QANet+CBBC+$v_1$& 79.37& 87.07& 74.91& 78.68& 62.69& 74.72& 61.40 \\
            QANet+CBBC+$v_2$& 79.50& 87.62& 74.94& 78.05& 61.97& 74.72& 61.12 \\
            QANet+CBBC+$v_3$& 80.97& 88.34& 76.52& 79.93& 62.86& 75.77& 62.52 \\
            QANet+CBBC+$v_4$& 81.54& 88.48& 77.89& 80.67& 63.79& 77.21& 63.83 \\
            QANet+CBBC+$\mathcal{V}_{corr}$& 83.75& 89.18& 80.04& 82.68& 65.35& 78.20& 64.53 \\
            \hline
            \textbf{QANet+CBBC+$\mathcal{V}$}& \textbf{85.13}& \textbf{90.46}& \textbf{82.40}& \textbf{85.08}& \textbf{66.81}& \textbf{78.88}& \textbf{66.66} \\
            \bottomrule
        \end{tabular}
     }
    \caption{Quantitative results of QANet backbone on four benchmark datasets. The best results are highlighted in \textbf{bold}.}
    \vspace{-4mm}
    \label{tab:comparisons-qanet}
\end{table*}

\section{Additional Diagrams of Samples' Capability-specific Value}
\label{sec:additional-diagrams}

The distribution diagrams of $v_2$ and $v_3$ are shown in Figure~\ref{fig:v23-value-distribution}.
They present a conclusion consistent with $v_1$ and $v_4$ discussed in Section~\ref{sec:relation-analysis}.
That is, among all the bins of $v_i$, the frequency of prediction results within the intermediate range ($0.4\sim0.6$) are similar ($ \approx 50\%$).
Furthermore, as the $v_i$ increases, the frequency of prediction results within a low range ($0.0\sim0.2$) also increases, while the one of a high range ($0.8\sim1.0$) decreases.

In addition to the distribution of model performance over different ranges of $v_i$, the mean value and standard deviation of model performance (in $F_{logits}$) over them and their subclasses are illustrated in Figure~\ref{fig:metrics-mean}.
As shown in the figure, it qualitatively shows the relatively strong correlation between the model performance and each capability-specific score.

\section{Additional Experiments Using Other Models}
\label{sec:additional-models}
In addition to the Transformer-based MRC model, we also perform ablation analysis using the following systems to further verify the effectiveness of our proposed assessment framework, whose training setting is consistent with that of our BERT-based MRC model.
\begin{itemize}
    \item R-Net~\cite{wang2017gated} matches the question and passage with gated recurrent neural networks (RNNs) to obtain the question-aware passage representation and employs the pointer networks to locate the positions of answer span from the passages.
    
    \item QANet~\cite{yu2018qanet} encodes the local and global interactions with the convolution and self-attention, respectively.
    It achieves higher training efficiency while obtaining the equivalent accuracy to the recurrent models.
\end{itemize}

The quantitative results of R-Net and QANet are summarized in Table~\ref{tab:comparisons-r-net} and Table~\ref{tab:comparisons-qanet}, respectively.
Our assessment framework also provides a significant performance improvement to these weaker backbones than BERT, such as RNN-based R-Net and convolution-based QANet.

\section{Additional Experiments Using Other Pipelines}
\label{sec:additional-pipelines}
To further verify the effectiveness of our proposed MRC competency assessment framework and reveal more available application scenarios for it, we embed it into the active learning pipeline besides curriculum learning.

\begin{figure}[!t]
	\centering
	
	\begin{subfigure}{1.0\columnwidth}
		\includegraphics[width=\textwidth]{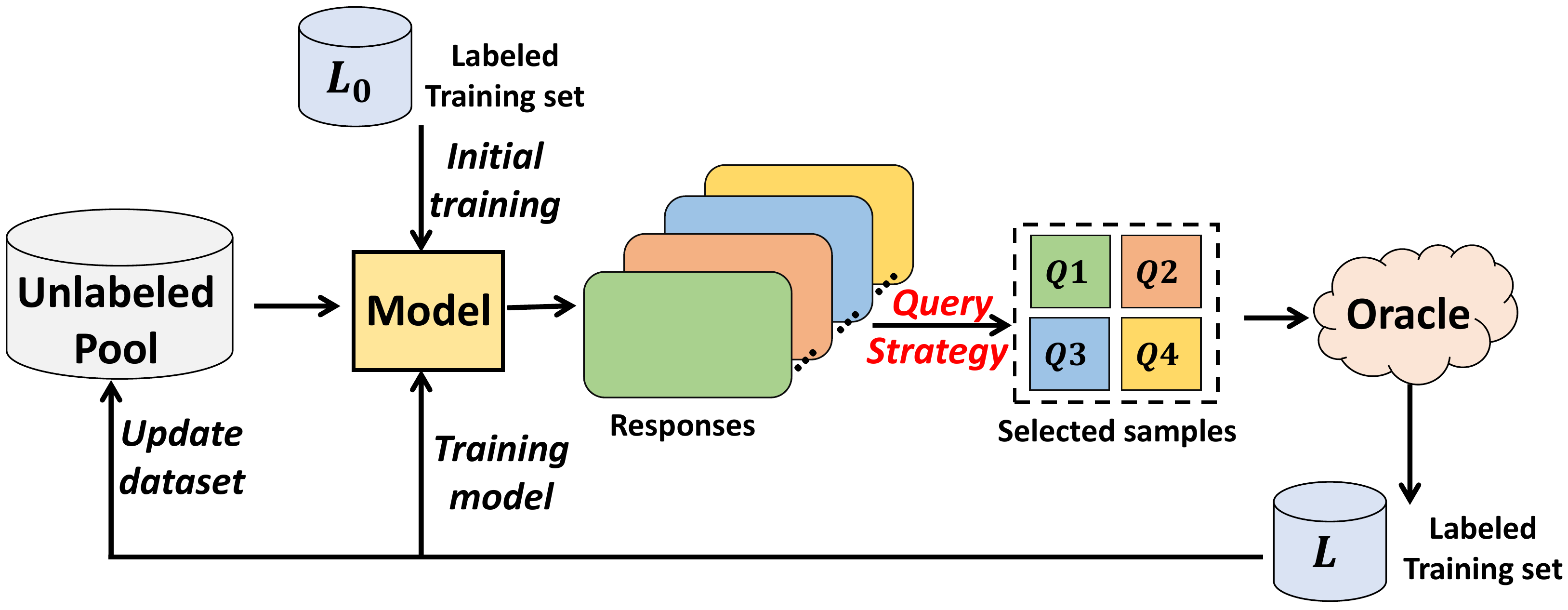}
		\subcaption{Pipeline of the pool-based active learning.}
		\label{fig:activelearning-pipeline}
	\end{subfigure}
    \\
	\begin{subfigure}{1.0\columnwidth}
		\includegraphics[width=\textwidth]{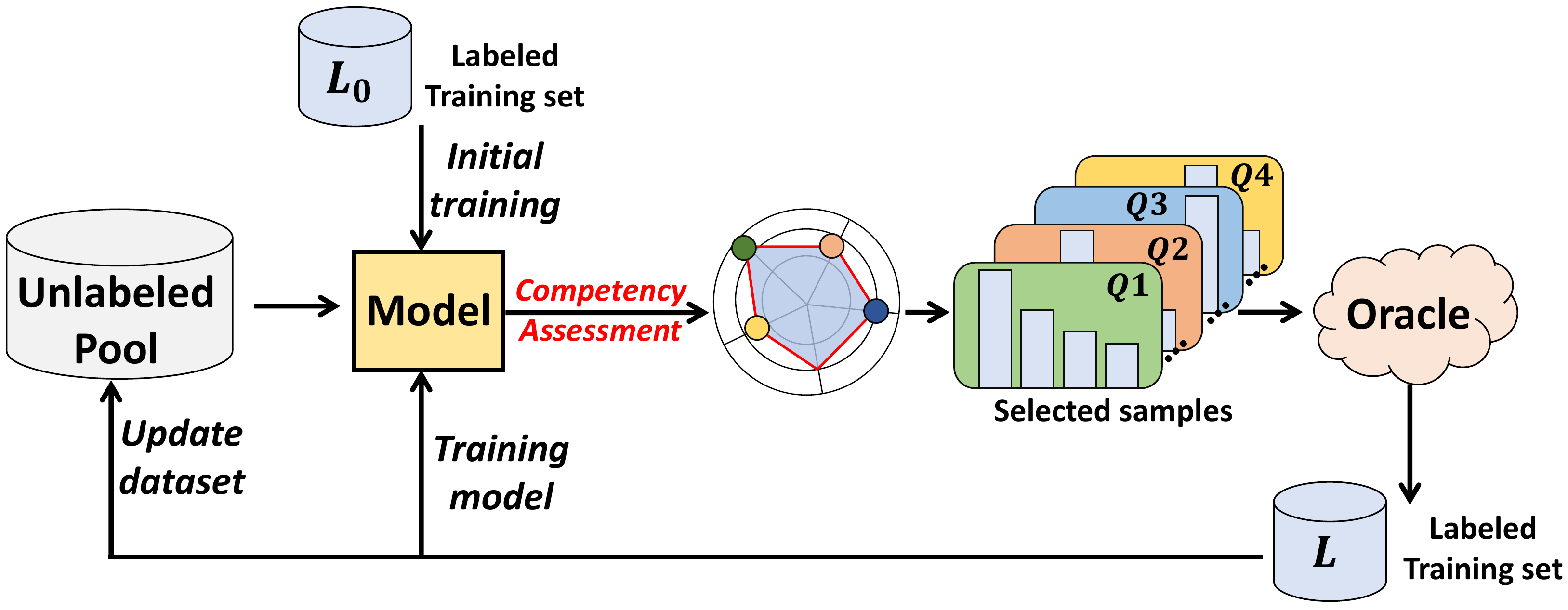}
		\subcaption{Pipeline of CBBC-guided active learning.}
		\label{fig:cbbc-activelearning-pipeline}
	\end{subfigure}

	\caption{Architecture comparison between the typical pool-based active learning pipeline and our CBBC-guided pipeline.}
\end{figure}

\begin{figure}[!t]
    \includegraphics[width=0.9\columnwidth]{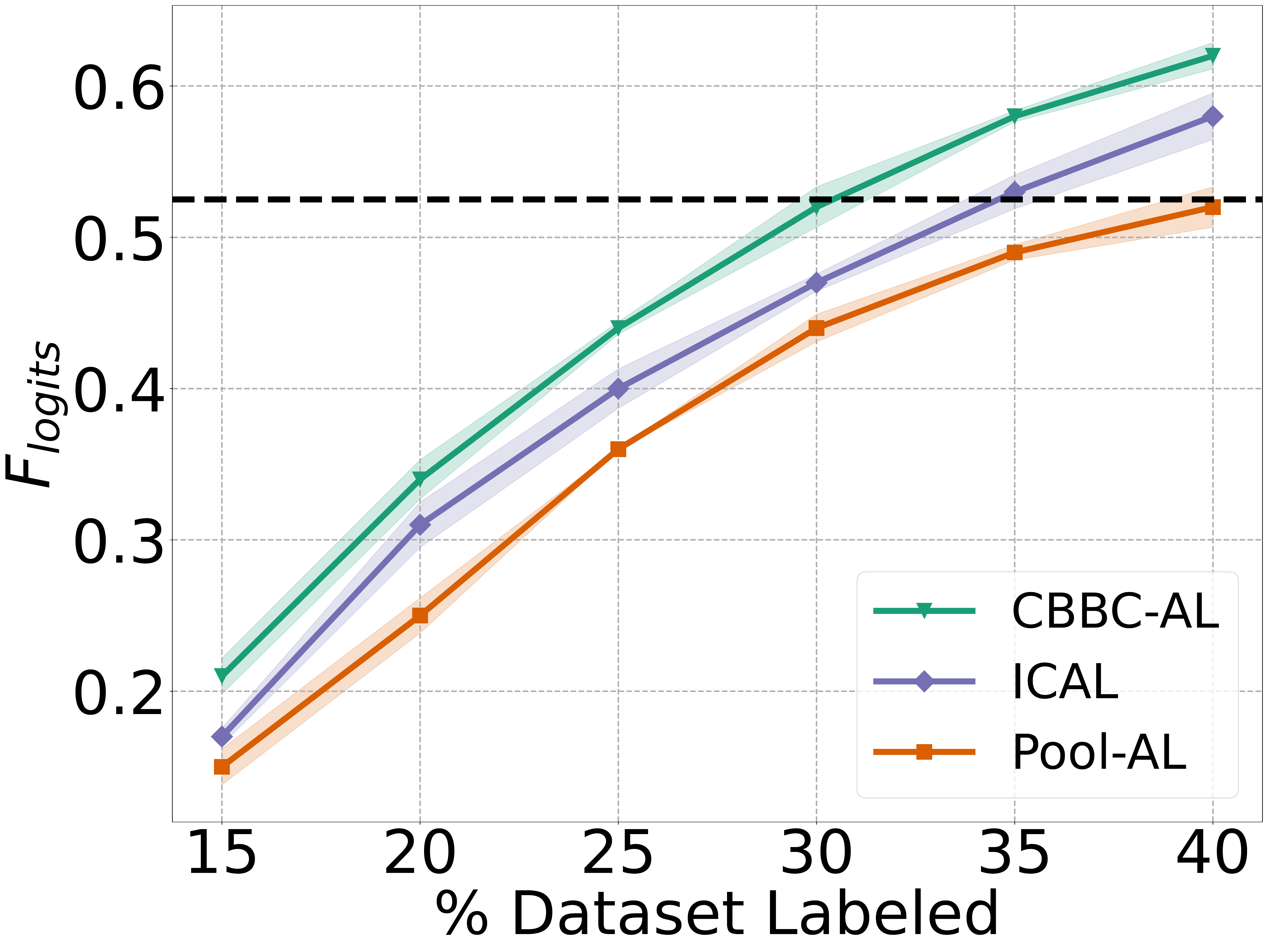}
    \caption{Performance comparison under active learning pipeline on the HotpotQA dev split.
    The solid lines indicate the results averaged over five trials, and shadows represent the standard deviation.
    The dotted line at the top represents the performance of the BERT-based MRC model with the whole training set labeled.}
    \label{fig:activelearning-results}
\end{figure}

\noindent
\textbf{CBBC in active learning.}
Given the training state of a model, active learning aims to select the most valuable samples from the unlabeled dataset and hand it over to the oracle (\eg human annotator) for labeling so as to reduce the cost of labeling as much as possible while still maintaining performance.
Take the most common pool-based active learning~\cite{lewis1994sequential, gal2016dropout} as an example, which queries the best sample based on the confidence evaluation and ranking of the entire dataset.
This query strategy is usually implemented by the uncertainty-based sampling~\cite{ebrahimi2019uncertainty, gal2017deep, houlsby2011bayesian, kirsch2019batchbald} and distribution-based sampling~\cite{pinsler2019bayesian, wei2015submodularity}.
Compared to the original pool-based active learning (shown in Figure~\ref{fig:activelearning-pipeline}), CBBC-guided active learning (shown in Figure~\ref{fig:cbbc-activelearning-pipeline}) provides a novel and interpretable query strategy by assessing the model in a 4-dimensional capability.
In addition to the pool-based baseline, we employ a more recent active learning pipeline ICAL~\cite{gao2020consistency} as a comparison, which selects samples with the high inconsistency of predictions over a set of data augmentations.

We employ BERT as our MRC backbone.
In each active learning cycle, we continue to train the MRC model by adding $5\%$ labeled data points by simulating the oracle annotating process.
The initial training set is randomly sampled from the HotpotQA train split and follows \citet{gao2020consistency} for the setting of initial training set size and active learning budget.

Figure~\ref{fig:activelearning-results} illustrates the results of different methods at each active learning cycle qualitatively.
Our CBBC-guided active learning (denoted as ``CBBC-AL'') achieve a higher MRC performance than the pool-based active learning (denoted as ``Pool-AL'') and ICAL from the start of training, demonstrating that the assessment of MRC model capability derived by our CBBC is reasonable and can also make a substantial difference to active learning beside curriculum learning.
When using only $35\%$ labeled samples, our CBBC-guided active learning outperforms the baseline model normally trained on the entire dataset by a considerable margin.

\end{document}